\documentclass{article}

\usepackage{microtype}
\usepackage{graphicx}
\usepackage{subfigure}
\usepackage{booktabs} 

\usepackage{hyperref}

\usepackage[accepted]{icml2024}

\usepackage{amsmath}
\usepackage{amssymb}
\usepackage{mathtools}
\usepackage{amsthm}
\usepackage{multirow}
\usepackage{enumitem}
\usepackage{subfigure}
\usepackage{bbm}
\usepackage{alphalph}
\usepackage{bm}

\usepackage[capitalize,noabbrev]{cleveref}

\theoremstyle{plain}
\newtheorem{theorem}{Theorem}[section]

\theoremstyle{definition}
\newtheorem{definition}[theorem]{Definition}
\newtheorem{assumption}[theorem]{Assumption}
\newtheorem{observation}[theorem]{Observation}
\theoremstyle{remark}

\usepackage[textsize=tiny]{todonotes}

\icmltitlerunning{Submission and Formatting Instructions for ICML 2024}

\begin{document}

\twocolumn[
\icmltitle{Dataset Clustering for Improved Offline Policy Learning}




\begin{icmlauthorlist}
\icmlauthor{Qiang}{1}
\icmlauthor{Yixin Deng}{1}
\icmlauthor{Francisco Roldan Sanchez}{2,3}
\icmlauthor{Keru Wang}{1}
\icmlauthor{Kevin McGuinness}{2,3}\\
\icmlauthor{Noel O'Connor}{2,3}
\icmlauthor{Stephen J. Redmond}{1,3}
\end{icmlauthorlist}

\icmlaffiliation{1}{University College Dublin, Ireland}
\icmlaffiliation{2}{Dublin City University, Ireland}
\icmlaffiliation{3}{Insight SFI Research Centre for Data Analytics, Ireland}

\icmlcorrespondingauthor{Stephen J. Redmond}{stephen.redmond@ucd.ie}

\icmlkeywords{Machine Learning, ICML}

\vskip 0.3in
]



\printAffiliationsAndNotice{\icmlEqualContribution} 

\begin{abstract}
Offline policy learning aims to discover decision-making policies from previously-collected datasets without additional online interactions with the environment. As the training dataset is fixed, its quality becomes a crucial determining factor in the performance of the learned policy. This paper studies a dataset characteristic that we refer to as \emph{multi-behavior}, indicating that the dataset is collected using multiple policies that exhibit distinct behaviors. In contrast, a \emph{uni-behavior} dataset would be collected solely using one policy. We observed that policies learned from a uni-behavior dataset typically outperform those learned from multi-behavior datasets, despite the uni-behavior dataset having fewer examples and less diversity. Therefore, we propose a behavior-aware deep clustering approach that partitions multi-behavior datasets into several uni-behavior subsets, thereby benefiting downstream policy learning. Our approach is flexible and effective; it can adaptively estimate the number of clusters while demonstrating high clustering accuracy, achieving an average Adjusted Rand Index of $0.987$ across various continuous control task datasets. Finally, we present improved policy learning examples using dataset clustering and discuss several potential scenarios where our approach might benefit the offline policy learning community. 
\end{abstract}

\section{Introduction} \label{sec:intro}
While deep reinforcement learning (DRL) has demonstrated the ability to surpass traditional control methods in various tasks \cite{atari,rrc2021-solving, rrc2021-dexterous}, its deployment in real-world applications remains limited. This limitation arises because DRL is thought of as a paradigm for online learning, usually requiring a large number of interactions with the environment, which can be costly and risky in the real world. Offline policy learning, which includes offline DRL \cite{bcq} and offline deep imitation learning (DIL) \cite{dil}, aims to identify a policy using a static dataset, eliminating the need for online interactions, and thus presenting a promising alternative to standard DRL. Offline policy learning is highly sample-efficient, enabling a single dataset to be utilized for multiple policy learning attempts.

Nonetheless, offline policy learning imposes a constraint, whereby policy learning is tightly bound to the static dataset. In this paper, we examine a dataset type that we refer to as \emph{multi-behavior}. This type of dataset is collected using various different policies, potentially resulting in a complex data distribution. Such datasets can arise in several real-world situations; for instance, where demonstration data comes from different humans or scripted policies exhibiting distinct skill levels \cite{multi-experts-1, multi-experts-2}. Conversely, we refer the dataset collected using only one policy as \emph{uni-behavior}.


In Appendix~\ref{append:multi-vs-uni-results}, we perform experiments to compare policies trained from both multi-behavior and uni-behavior datasets, respectively. We have incorporated several different tasks from the D4RL benchmark \cite{d4rl}, along with various state-of-the-art learning algorithms \cite{iql, td3bc, crr, bc}. The results suggest that using uni-behavior datasets results in quicker and more stable policy convergence, as well as superior policy performance compared to policies learned using multi-behavior datasets, even though the uni-behavior datasets are much smaller in size and have less exploration diversity compared to the multi-behavior datasets. This could be attributed to the fact that the various policy behaviors contained in a multi-behavior dataset may lead to multimodal action distributions when conditioned on the states \cite{single-vs-multi-distributions1, single-vs-multi-distributions2}. This may be especially harmful when performing supervised offline policy learning, such as behavioral cloning \cite{bc_tutorial}.

Based on the results shown above, we are aware that using the same dataset in different ways can lead to markedly different policy learning outcomes. Interestingly, in some instances, the policy learned from a subset of the training dataset may outperform that learned from the entire dataset. Therefore, in this paper, we aim to unlock the potential of the multi-behavior dataset by clustering it into a group of learning-friendly uni-behavior subsets. It is worth noting that the primary focus of this paper is not on proposing any specific policy learning algorithm, but rather on presenting a data processing approach that can broadly improve the performance of policy learning algorithms in multi-behavior scenarios. The main contributions of this paper include: (\textit{i}) We identified an intrinsic feature in the action trajectories of policy learning data and utilized it to propose a simple clustering approach. This feature enables classical K-means clustering to cluster the dataset effectively; (\textit{ii}) We developed a behavior-aware deep clustering method that performs clustering by learning and contrasting various behaviors within the multi-behavior dataset. It surpasses K-means and operates without the need to pre-define the exact number of clusters; and (\textit{iii}) Our approach has been tested on $10$ different multi-behaviour datasets from $10$ respective tasks, with each multi-behavior dataset containing $6$ uni-behavior subsets. These datasets encompass both simulated and real-world environments. We have open-sourced all the datasets and code to serve as a benchmark of multi-behavior clustering and policy learning for the community.

\section{Related work} \label{sec-relatedworks}
Clustering is an unsupervised method that aims to partition data into groups based on similarity, with the goal of optimizing intra-cluster similarity while maximizing inter-cluster dissimilarity \cite{clu-review}. The K-means algorithm \cite{kmeans}, widely employed in clustering tasks, groups data points into clusters around $K$ centroids with the objective of minimizing the Euclidean distances between points and their nearest centroid. However, due to the Euclidean distance metric, it tends to favor spherical clusters and may not effectively handle clusters with complex shapes \cite{kmeans-sph}. Furthermore, K-means requires a predefined cluster count ($K$), which may not be known in many cases. Density-based clustering methods \cite{dbscan, optics}, unlike K-means, can identify clusters of various shapes without predefined cluster numbers. They classify points as a cluster if at least $\phi$ points are within an  $\epsilon$  radius in Euclidean space \cite{dbscan, optics}. However, their effectiveness depends significantly on the selection of $\phi$ and $\epsilon$. More recently, deep clustering \cite{deep-clu-review} has been proposed as a fusion of deep learning with traditional algorithms, as mentioned above. It excels at handling high-dimensional data; however, it inherits the shortcomings of the traditional clustering methods on which it is based, and suffers from poor interpretability and high computational complexity.

K-means clustering has been used in several studies to improve policy learning. \citet{clue} employs K-means to group offline policy learning data into $100$ distinct clusters based on state-action transitions. Each cluster is anticipated to represent a unique behavior, enabling the downstream DRL agent to acquire diverse skills from them. A similar clustering approach is employed in \cite{reducing} to improve offline policy learning from imbalanced multi-behavior datasets. In this approach, K-means is used to group transitions with various behavior patterns into clusters; subsequently, greater weights are allocated to expert clusters with small size for policy learning. Moreover, \citet{cer} employs K-means to group explored experience into various clusters based on transitions. They subsequently create a conditional probability density function to facilitate the fair replay of these clusters in DRL, hence mitigating the neglect of certain transitions in the buffer, thereby improving exploration efficiency. Likewise, \citet{clu-rl} uses K-means on state data to improve exploration efficiency, forming clusters to represent different environmental areas and adapting different exploration strategies. 

Learning-based clustering methods are also utilized to improve policy learning. Classifiers are trained by contrasting expert behaviors with manually-created weak behaviors in \cite{pubc, rrc2022}. Subsequently, classifiers are used to group the dataset into two clusters, with one derived from the expert, which is then utilized for policy learning. \citet{ood-clu} introduces a clustering method based on contrastive learning to group data into clusters with varying environmental dynamics. The objective of the clustering algorithm is to mitigate interference between different dynamics within the same dataset.

Our proposed clustering approach surpasses previous methods in efficacy and flexibility, functioning without predefined cluster numbers and accommodating diverse cluster shapes. It minimizes dependence on parameter adjustments, offering interpretable results. To our knowledge, this paper is the first to explicitly cluster multi-behavior data into uni-behavior subsets and discuss the advantage of learning from such datasets. 

\section{Preliminaries}
\subsection{Notations and settings}
The DRL environment can be formulated as the Markov decision process and is defined by the tuple $\left \langle \mathcal{S}, \mathcal{A}, T, r, \rho_{0}, \gamma  \right \rangle $, where: $\mathcal{S}$ denotes the state space, $\mathcal{A}$ denotes the action space, $T(\textbf{s'}\mid \textbf{s}, \textbf{a})$ denotes the environmental dynamic transition ($\textbf{s'}$ is the next state after applying action $\textbf{a}$ at $\textbf{s}$), $r(\textbf{s}, \textbf{a}): \mathcal{S} \times \mathcal{A} \mapsto \mathbb{R}$ is the reward function, $\rho_{0}$ denotes the initial states, and $\gamma\in (0,1]$ denotes the discount factor. In this context, \( \textbf{s} \) and \( \textbf{a} \) are in bold to show they are multi-dimensional vectors in state and action spaces.

The dataset in our work, denoted as $\mathcal{D}$, is collected from interactions between the pre-trained policy $\pi$ and the environment. $\mathcal{D}$ comprises $n$ trajectories and can be written as $\mathcal{D} = \{\tau_{1}, \tau_{2}, \ldots, \tau_{n}\}$.  Each trajectory contains transitions, and the $j^{th}$ trajectory with $k$ transitions can be written as: $\tau_{j} = \left\{ (\mathbf{s}_0, \mathbf{a}_0, r_0, \Omega_0), (\mathbf{s}_1, \mathbf{a}_1, r_1, \Omega_1), \ldots, (\mathbf{s}_k, \mathbf{a}_k, r_k, \Omega_k) \right\}$. Here, $\textbf{s}$ represents the state vector, $\textbf{a}$ the action vector, $r$ the reward value, and $\Omega$ is a Boolean value indicating whether the transition is the terminal point of a trajectory.

In our work, the multi-behavior datasets are created by combining multiple uni-behavior datasets collected using different policies. Each policy exhibits a unique stationary behavioral pattern, which can be understood intuitively: the actions, when conditioned on the state under different policies, should vary. However, considering that actions can involve long-horizon decision sequences and be stochastic when interacting with the environment, even for different policies, the same actions may be observed given the same state. Therefore, we define different policies to be:

\begin{definition}
Given two policies, \(\pi_{q}\) and \(\pi_{p}\), if they exhibit different behaviors, then:
\begin{equation}
\exists \textbf{s} \in S: \overline{\pi_{q}(\textbf{s})} \neq \overline{\pi_{p}(\textbf{s})},
\end{equation}
where \(\overline{\pi(\textbf{s})}\) represents the expected action at state \( \textbf{s} \) according to policy \(\pi\). If the action space is discrete, it can be calculated by $\overline{\pi(\textbf{s})} = \sum_{\textbf{a} \in \mathcal{A}} \textbf{a} \cdot \pi(\textbf{a} | \textbf{s})$. If the action space is continuous, it can be calculated by: $\overline{\pi(\textbf{s})} = \int_{\mathcal{A}} \textbf{a} \cdot \pi(\textbf{a} | \textbf{s}) \, d\textbf{a} $.
\end{definition}

\subsection{Tasks and datasets} \label{subsec:task-and-dsets}
Our datasets are collected from $10$ different tasks across $3$ benchmark suites. Below, we offer a brief introduction to these tasks and datasets. The detailed configuration of each dataset is reported in Appendix \ref{append:dset-compo}. 

\subsubsection{Tasks}
\paragraph{Locomotion} Our work involves $5$ locomotion tasks from OpenAI Gym \cite{openai-gym}: \emph{HalfCheetah}, \emph{Hopper}, \emph{Humanoid}, \emph{Walker2d}, and \emph{Ant}. These tasks are widely used in the field of offline policy learning as benchmarks \cite{d4rl, iql, td3bc, crr}. These locomotion tasks necessitate that the agent coordinates the joints of geometric bodies to counteract the force of gravity in the environment, maintain balance, and advance forward.
\vspace{-4mm}
\paragraph{Robotic hand manipulation} Our work involves 3 robotic hand manipulation tasks from the DAPG project \cite{dapg}: \emph{Hammer}, \emph{Door}, and \emph{Pen}. These tasks involve controlling the robotic hand to accomplish specific objectives: picking up a hammer to drive a nail into a board, undoing a latch and swinging a door open, and repositioning a pen to match the orientation of a target.
\vspace{-4mm}
\paragraph{Tri-finger robotic manipulation} Our work involves $2$ real-world physical dexterous manipulation tasks using a robotic system composed of three fingers \cite{2022rrc}: \emph{Push} and \emph{Lift}. These tasks require the robot to manipulate a cube in two ways: pushing the cube towards a 2D target position on the floor, and lifting or/and rotating the cube to achieve the 3D target position and orientation.

\subsubsection{Collocating datasets} 
For the locomotion and robotic hand manipulation environments, we collected $6$ uni-behavior datasets for each task and implemented the following settings to make the merged multi-behavior datasets more realistic and challenging for clustering: \textbf{\textit{(i)} Uneven data quality}: The policies employed for dataset collection display varying skill levels, wherein certain expert policies may yield more consistent actions, while weaker policies might result in actions exhibiting greater randomness; \textbf{\textit{(ii)} Similar behavioural patterns}: We incorporate similar behavior patterns into the datasets to add complexity to the clustering task. To achieve this, we specify that certain data collection policies must closely align with task-solving skill levels in terms of rewards. As a result, each multi-behavior dataset includes two expert-level policies, two intermediate-level policies, and two weaker policies. \textbf{\textit{(iii)} Identical policy output function}: We stipulated that policies employed for the same task must be trained using the same algorithm and feature the same policy output function. This measure is taken  to increase the complexity of the clustering task, as it reduces the likelihood of clustering tendencies arising from biases and inconsistencies stemming from different output functions. We used different random seeds to ensure that the trained policy exhibits distinct behavioral patterns. \textbf{\textit{(iv)} Diversity of results}: Our work involved several different types of policy learning algorithms, including PPO \cite{ppo}, TD3 \cite{td3}, and SAC \cite{sac}. These encompass both deterministic and stochastic algorithms, aimed at ensuring that our method can adapt to datasets with diverse characteristics.

The multi-behaviour datasets for tri-finger robot manipulation tasks were not collected in this work but obtained from an open-source project in \cite{2022rrc}. Each multi-behaviour dataset was collected using two different policies: one being an expert, and the other being a weaker policy. The policies for each task are from the same training trial, with expert policies coming from later checkpoints and weak policies from earlier checkpoints, resulting in somewhat similar behavioral patterns but differing skill levels.

\section{Exploring the tendency of action data to cluster}
\subsection{Analyzing the action data} \label{subsec:analy-action-data}
Our clustering method starts with the analysis of the action data. While action data alone cannot fully reveal the decision-making processes of a policy without taking the external environment into account, on the average it does provide a direct reflection of the \emph{intrinsic behavioral features} of a policy. For example, different humans performing physical tasks may display distinctive patterns of force application, speed, and pose when performing daily tasks. These intrinsic behavioral features of humans are determined by physiological and psychological differences between individuals and have relatively low correlation with the environment. Sometimes, these intrinsic behavioral features can be used to quickly identify different people; hence, actions can potentially be used to cluster multi-behavior datasets. By examining the action data, we make the following observations:
\begin{observation} \label{observation: 1}
\textit{
The mean of the lower percentile portion of pairwise Euclidean distances, calculated within actions from the same behavioral policy, is smaller than that calculated between actions derived from different policies. This can be expressed as:
\begin{equation} \label{eq:obs1}
\resizebox{\columnwidth}{!} {%
$\mathbb{E}_{\bm{a}_{1}, \bm{a}_{2} \sim \mathcal{D}^{\mathcal{A}}_{\pi_q}} [\mathcal{P_{\downarrow }}(d(\bm{a}_{1}, \bm{a}_{2}))] < \mathbb{E}_{\bm{a}_{1} \sim \mathcal{D}^{\mathcal{A}}_{\pi_q}, \bm{a}_{2} \sim \mathcal{D}^{\mathcal{A}}_{\pi_p}} [\mathcal{P_{\downarrow }}(d(\bm{a}_{1}, \bm{a}_{2}))],$%
}
\end{equation}
where $\mathcal{D}^{\mathcal{A}}_{\pi_q}$ and $\mathcal{D}^{\mathcal{A}}_{\pi_p}$ represent the action sets generated by policy $\pi_{q}$ and  $\pi_{p}$ ($\pi_{q}$ $\neq$ $\pi_{p}$). The function $d$ represents the calculation of Euclidean distance. $\mathcal{P_{\downarrow }}$  represents the process of filtering  by retaining a small percentile of the smallest values of  $d(a_{1}, a_{2})$ for each pair of tuples $(a_{1}, a_{2})$, as per Equation (\ref{eq:obs1}). This observation is further validated in the Appendix~\ref{append:subsec:obs1}. 
}
\end{observation}

Observation \ref{observation: 1} informs us that action data generated by different policies each have relatively concentrated regions in the action space; i.e., high-density areas. These areas tend to cluster into distinct groups, each with a different group centroid, i.e., $\mu _{\pi_{q}} \neq \mu _{\pi_{p}}$, where $\pi_{q}\ne \pi_{p}$. This also supports our above statement, which is that actions, as intrinsic behavioral features, can sometimes reflect distinctions between various policies.

\subsection{Long-horizon clustering feature} \label{subsec:action-rep}
Despite the above observation, directly using action data for clustering yields unsatisfactory outcomes (refer to Table \ref{table:cluster-results}). The reason is that action data from different policies have significantly overlapping distributions (refer to the results in Appendix~\ref{append:subsec:obs1}); this is incompatible with effective clustering of an entire dataset. This could be attributed to the fact that actions typically involve long-horizon decision sequence in control tasks. As a result, a single action is insufficient to effectively encapsulate the behavioral patterns of a policy. Hence, we propose a long-horizon feature that summarizes the action sequence to more effectively reflect the behavioral patterns of policies; this is achieved by averaging the time series action trajectories, and we refer to it as the \textit{temporal-averaged action trajectory} (TAAT), performed by:
\vspace{-0.75mm}
\begin{equation}
\label{eq: action-rep}
\bm{\overline{\tau^{\mathcal{A}}}} = \textstyle \frac{1}{T} \sum_{t=1}^{T} \bm{a}_{t},
\end{equation}
where $T$ represents the number of actions in each trajectory, and $i$ represents the $i^{th}$ component in the action vector.  $\bm{\overline{\tau^{\mathcal{A}}}}$ is in bold to show it is multi-dimensional vectors. This operation can be understood as the application of the Weak Law of Large Numbers (WLLN). However, WLLN requires that the actions are independent and identically distributed, but consecutive actions may not be entirely independent. Therefore, we further substantiate the above point through experiments in Appendix \ref{append:prove-eq}. 

\begin{figure}[h]
    \centering
        \subfigure[Without TAAT]{\includegraphics[width=0.48\linewidth]{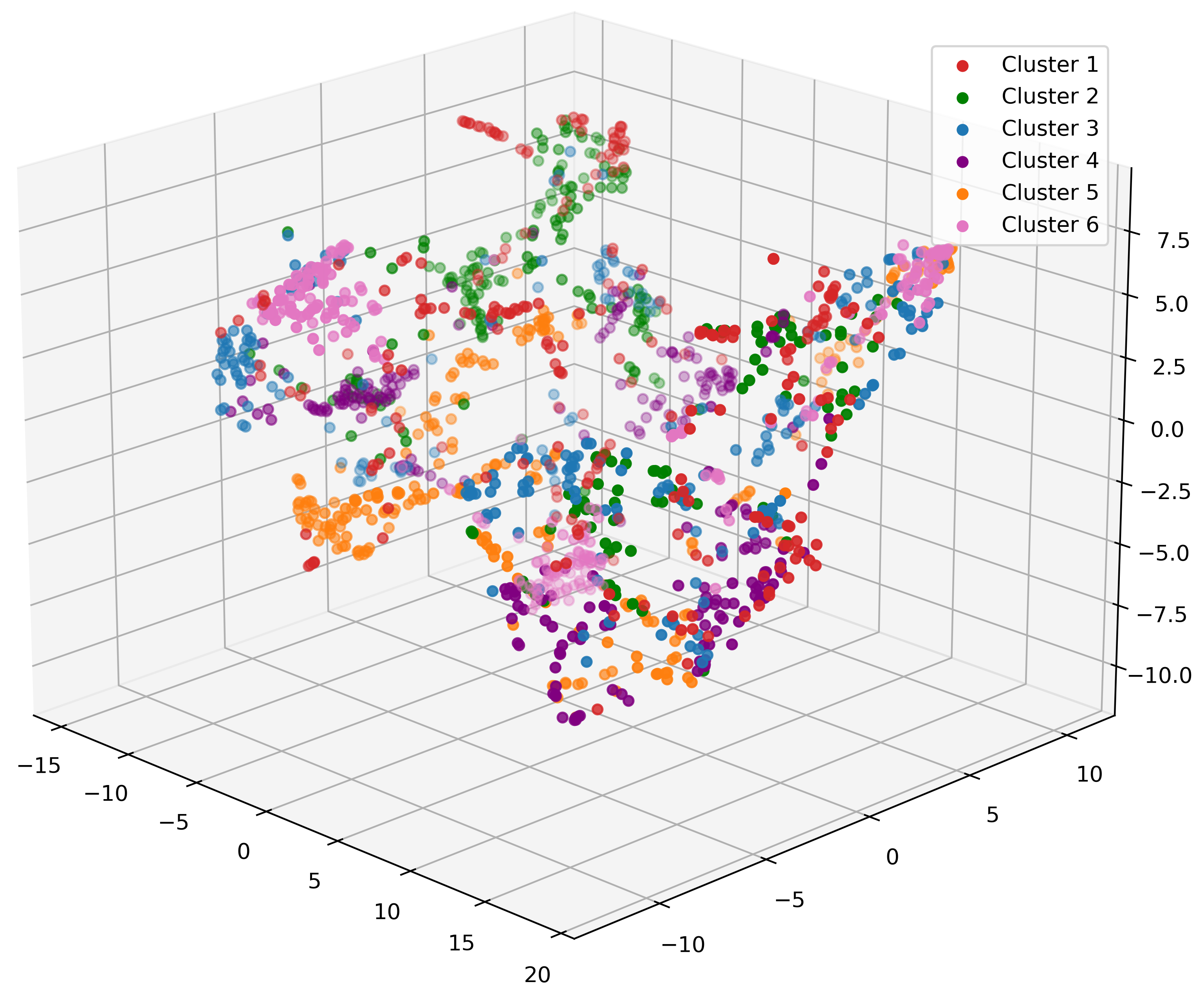}}
        \hspace{1mm}
        \subfigure[With TAAT]{\includegraphics[width=0.48\linewidth]{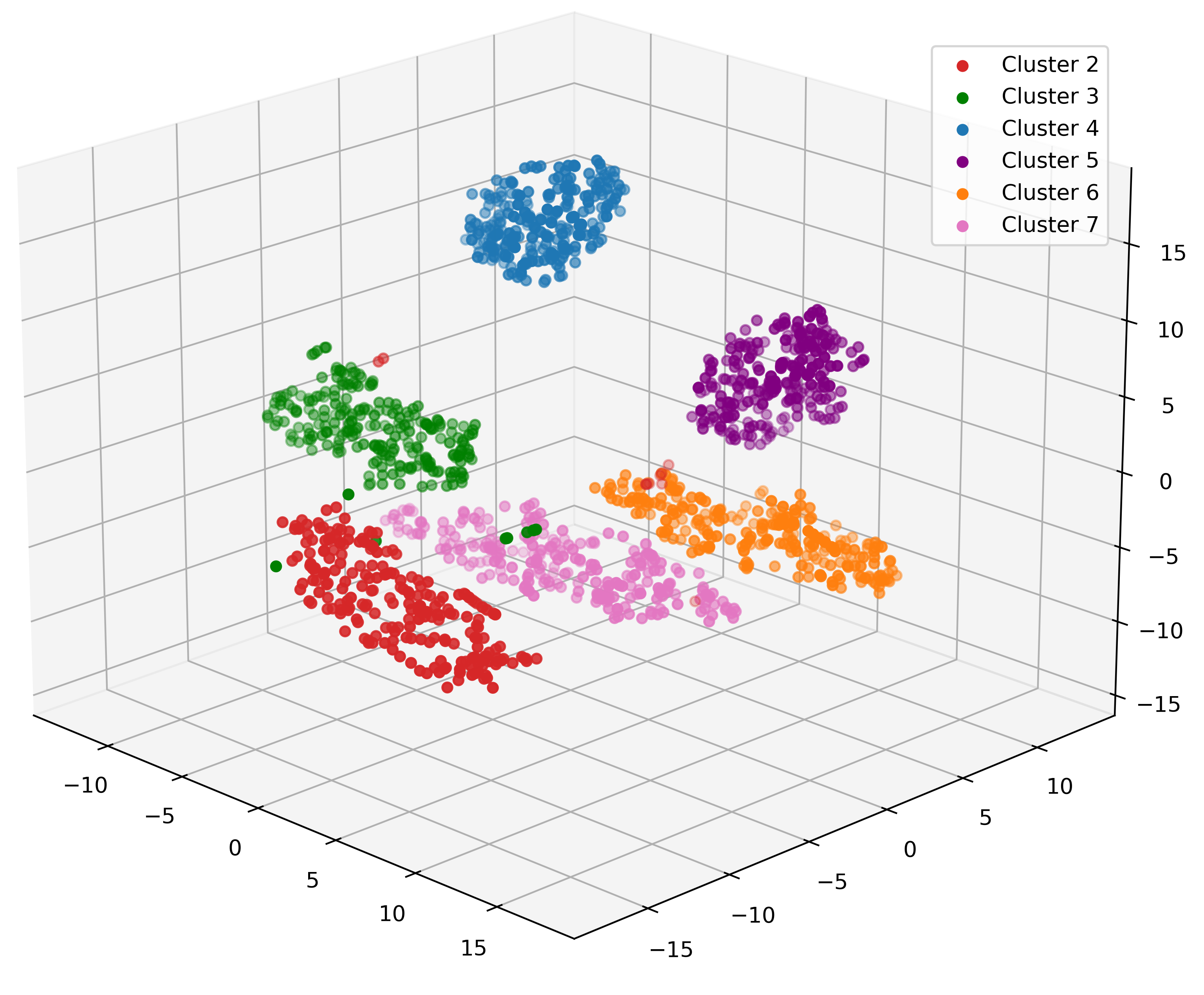}}
    \vspace{-2mm}
    \caption{Illustration showing the distribution of data points in Euclidean space, both with and without TAAT, on the Halfcheetah task's multi-behavior dataset. For the left plot, we randomly sample 3,000 action vectors from the entire dataset. For the right plot, we average each trajectory to obtain 3,000 TAAT vectors. We used t-SNE \cite{t-SNE} to reduce the dimensionality of the vectors to 3D for visualization.}
    \label{fig:vis-with-without-rep} 
\vspace{-1mm}
\end{figure}

We provide a visualization example of the distributions of the data points in Euclidean space in Figure \ref{fig:vis-with-without-rep} for comparison, both with and without the use of TAAT. Here, we used t-SNE \cite{t-SNE} to reduce the dimensionality of the vectors to 3D for visualization. As can be observed, utilizing TAAT can lead to a noticeable clustering trend among the action trajectories generated by different policies in the Euclidean space. We offer quantitative evaluations of TAAT in Table \ref{table:cluster-results} and Table \ref{table:validate-eq9} to illustrate the effectiveness of TAAT approach. The results demonstrate that our TAAT significantly enhances the performance of K-means in clustering multi-behavior datasets compared to using K-means without TAAT.

\begin{figure*}[!htbp]
\begin{center}
    \centerline{\includegraphics[width=0.995\linewidth]{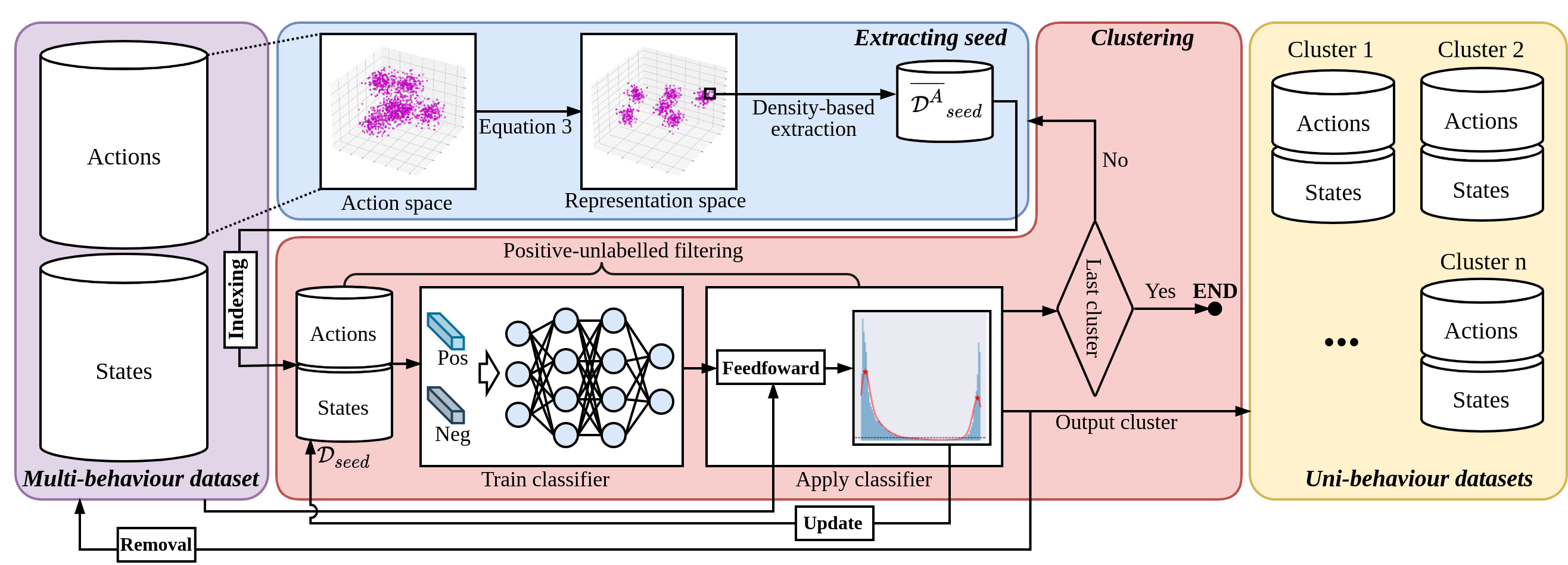}}
    \caption{Illustration of the flowchart for our iterative clustering algorithm. The purple box contains the raw multi-behavior dataset. The blue box represents the extraction of uni-behavior seed subsets for training the subsequent positively-unlabelled (PU) filter. Notably, the blue box only extracts the action sequences from trajectories, and we need to retrieve the corresponding state sequence of the trajectory from the purple box based on the corresponding indexes. The red box represents one clustering iteration, including training and using the PU filter for clustering, updating the original multi-behavior dataset by removing the resulting cluster from it, as well as checking whether to terminate the clustering process at this stage. Finally, the yellow box shows the results of uni-behavior clusters.}
\label{fig:flowchart}  
\end{center}
\vskip -0.2in
\end{figure*}

\section{Behaviour-aware deep clustering} \label{sec:clu-main}
Although K-means can cluster TAAT to some extent, its practicality is limited by the requirement of the predefined cluster numbers. The elbow method, commonly employed to determine the optimal number of clusters for K-means, is not universally applicable to all clustering tasks \cite{kmean-elbow}. In our work, the elbow method proved ineffective across all tasks in identifying the appropriate clustering number, as detailed in Appendix \ref{append:kmeans-elbow}. Therefore, we propose a more flexible, behavior-aware clustering method that can effectively estimate the correct number of clusters. Unlike traditional methods that group data directly into a number of clusters, our approach involves iteratively filtering out clusters of trajectories where each cluster contains trajectories with similar behavioral patterns.

The iterative process is illustrated in Figure \ref{fig:flowchart}. In brief, we first extract a small subset of uni-behavior trajectories from the entire dataset as the initial seed. This extraction is based on identifying examples from a region with high density using the long-horizon TAAT proposed above. Secondly, a binary classifier is trained to learn the behavioral pattern in the seed set, and then it is used to filter out all trajectories with the same behavioral pattern as the seed from the entire dataset through a semi-supervised iterative process. The above process (selecting seeds, training classifiers, and deploying the classifier for filtering) continues iteratively until the termination condition is met. This condition indicates that the cluster currently being processed is the final one. We offer an algorithm in Appendix \ref{append:algorithm} to facilitate the understanding of this iterative process.

\subsection{Extracting uni-behaviour seed set} \label{subsec:extract-seed-set}
Observation \ref{observation: 1} indicates that a portion of the action data generated by the same policy tends to form a high-density region in space. After transformation through Equation \ref{eq: action-rep} the clustering of trajectories becomes tighter. This leads us to the following assumption:
\begin{assumption} \label{assumption: 2}
\textit{Given a multi-behaviour action set $\mathcal{D^{A}}$, which is transformed to a TAAT set $\overline{\mathcal{D^{A}}}$ using Equation \ref{eq: action-rep}. If we identify a subset \( G \) with the highest density in \( \overline{\mathcal{D}^\mathcal{A}} \), \( G \) may largely consist of uni-behavior data. We determine \( G \) by finding samples that possess the shortest mean pairwise distances, which can be expressed as:}
\begin{equation}
\overline{\mathcal{D}^{\mathcal{A}}}_{seed} = \underset{G \subseteq \overline{\mathcal{D^{A}}}}{\mathrm{argmin}} \left( \frac{1}{g} \sum_{\substack{\bm{\overline{\tau^{\mathcal{A}}_{q}}}, \bm{\overline{\tau^{\mathcal{A}}_{p}}} \in G \\ q \neq p}} d(\bm{\overline{\tau^{\mathcal{A}}_{q}}}, \bm{\overline{\tau^{\mathcal{A}}_{p}}}) \right),
\end{equation}
\textit{where $G$ contains $g$ trajectories. To obtain \(\overline{\mathcal{D}^\mathcal{A}}_{seed} \), theoretically, exhaustive search of all possible combinations of \( G \) yields the most accurate result. However, this approach is computationally impractical for large datasets. To address this, we employ the Monte Carlo search (MCS) method to approximate the optimal \( G \). Specifically, we randomly sample \( z \) subsets of size \( g \) from \( \overline{\mathcal{D}^\mathcal{A}} \) for each experiment, then calculate the mean pairwise distance within each of these \( z \) subsets, and finally select the subset with the smallest mean distance as \( \overline{\mathcal{D}^\mathcal{A}}_{seed}\). The effectiveness of this assumption is further validated through experiments in Appendix \ref{append:prove-assum2}. These experiments demonstrate that choosing \( z = 10^6 \) and \( g = 6 \) yields satisfactory results.}
\end{assumption}

Once \(\overline{\mathcal{D}^\mathcal{A}}_{seed} \) has been determined through the MCS method, we take a further step to optimize \( \overline{\mathcal{D}^\mathcal{A}}_{seed} \). This optimization aims to increase the size of \( \overline{\mathcal{D}^\mathcal{A}}_{seed} \), thus providing a larger dataset for the subsequent training of neural networks. We know that the obtained \(\overline{\mathcal{D}^\mathcal{A}}_{seed} \) is characterized by a higher density, hence it is likely located in a high-density area and such high-density areas are usually composed of data points from the same policy under the theoretical framework in \citep{dbscan}. Therefore, we consider the Euclidean center of \(\overline{\mathcal{D}^\mathcal{A}}_{seed} \) as its centroid and identify the \( g_2 \) nearest neighbors around this center to obtain a subset $G_2$, where \( g_2 \) can be a relatively large number. In our case, \( g_2 \) is set to $\sim2-6$\% of the total number of trajectories in the raw multi-behaviour dataset. We then take $G_2$ as the new \(\overline{\mathcal{D}^\mathcal{A}}_{seed} \). It should be noted that this method (updating the centroid and finding new neighbors) can still iterate, but we have found that it can converge to satisfactory results by iterating only once, as described above. Examining the results of $\overline{\mathcal{D}^\mathcal{A}}_{seed}$ allows us to easily obtain the complete seed dataset $\mathcal{D}_{seed}$ containing both states and actions through checking the corresponding indices.

\subsection{Positive-unlabelled filter}
In the above section, we identified a uni-behavior seed subset $\mathcal{D}_{seed}$. This section aims to filter out the remaining trajectories that exhibit the same behavioral pattern as observed in $\mathcal{D}_{seed}$ from the entire multi-behavior dataset $\mathcal{D}$. To accomplish this, we introduce a learning-based positive-unlabelled (PU) filter primarily based on the approach described in \citep{pubc}. PU learning setting refers to a specific learning scenario in semi-supervised learning where a subset (the seed set $\mathcal{D}_{seed}$ in our case) of the entire dataset (the multi-behavior dataset $\mathcal{D}$ in our case) is assigned positive labels, with the remaining data being unlabelled \cite{pu-setting}. The PU filter is essentially a binary classifier that is trained to distinguish between positive samples from $\mathcal{D}_{seed}$ and artificially generated negative samples. A semi-supervised learning paradigm is implemented to iteratively update $\mathcal{D}_{seed}$ until convergence. We will now introduce each key component of this PU filter.

\subsubsection{Generate training samples} \label{subsec:generate-samples}
The positive samples are the state-action pairs from $\mathcal{D}_{seed}$. Negative samples are state-action pairs that do not exist in the dataset but are generated by mixing states and actions from different data sources, such that the resulting state-action pairs are very different to the available labelled state-action pairs in $\mathcal{D}_{seed}$. Specifically, this involves: (1) combining states/actions randomly sampled from $\mathcal{D}_{seed}$ with actions/states randomly sampled from the unlabelled set ($\mathcal{D} - \mathcal{D}_{seed}$), (2) combining states/actions randomly sampled from the space with actions/states randomly sampled from either $\mathcal{D}_{seed}$ or the remaining unlabelled subset, and (3) generating novel state-action pairs by sampling from the state and action spaces.


\subsubsection{Filter structure}
The PU filter model is a binary classifier that takes the state-action pair as input with the goal if minimizing:
\begin{equation}
\label{eq:cross-entropy-loss}
\begin{split}
\theta^* &= \underset{\theta}{\text{argmin}} \left( \mathbb{E}_{(\bm{s},\bm{a})\sim\mathcal{D}_+} \left[ -\log(1 - \mathcal{F}(\bm{s},\bm{a}; \theta)) \right] \right. \\
&\quad \left. + \mathbb{E}_{(\bm{s},\bm{a})\sim\mathcal{D}_-} \left[ -\log \mathcal{F}(\bm{s},\bm{a}; \theta) \right] \right),
\end{split}
\end{equation}
where \(\theta\) denotes the parameters of the classifier $\mathcal{F}$, and $\mathcal{D}_+$  and \(\mathcal{D}_-\) denote positive and negative sample sets generated in Section \ref{subsec:generate-samples}. The activation function of the final layer is sigmoid, defined as $\operatorname{sigmoid}(x) = {1}/{\left(\textstyle 1 + e^{-x}\right)}$, allowing the outputs to be interpreted as probabilities. To further improve the robustness and accuracy, multiple classifiers are trained simultaneously using the bagging technique, and their individual decisions are then combined through weighted voting for a more comprehensive final outcome.

\subsubsection{Using the trained classifier to filter} \label{subsubsec:use-pu-filter}
The estimated probabilities of individual state-action pairs are aggregated over the trajectory: $\tau_{prob} = \frac{1}{T} \sum_{t=1}^{T} \mathcal{F}(\bm{s}_{t},\bm{a}_{t})$, where $T$ represents the number of transitions in the trajectory. $\tau_{prob}$ can be interpreted as the probability of one trajectory exhibiting the same behavior pattern as the trajectories in the seed subset. The next step involves converting these continuous probabilities into Boolean labels; i.e., filtering. A threshold $th_{prob}$ is required for this conversion: $\mathbbm{1}(\tau_{prob} > th_{prob})$, where $\mathbbm{1}(\cdot)$ represents an indicator function. In our work, an adaptive threshold is implemented to ensure more adaptive filtering rather than using a fixed threshold \cite{fixed-th}. Trajectories with behavior patterns similar to the seed subset are generally assigned higher probabilities by the classification and aggregation process described above, while the remaining trajectories obtain lower probabilities. This results in bimodality of the histogram of output probabilities (see Figure \ref{fig:adap-th}). Therefore, we conduct kernel density estimation (KDE) with a Gaussian kernel to model what is assumed are two overlapping distributions, and then identify the threshold by selecting the probability value at the maximum local-minimum point on the KDE curve. An example illustrating the selection of this threshold is provided in Appendix \ref{append:subsec-adaptive-th}. 

Through the above steps, additional trajectories that exhibit the same behavioral patterns observed in $\mathcal{D}_{seed}$ can be identified. These trajectories are then added to $\mathcal{D}_{seed}$, which is subsequently utilized for training the next classifier. The semi-supervised iteration process terminates when the membership in $\mathcal{D}_{seed}$ reaches convergence.

\subsection{Last clustering iteration check}\label{subsec:check-last-cluster}
Section \ref{subsubsec:use-pu-filter} mentions that the probability histogram output by the PU filter displays two distinct distributions. The larger distribution contains trajectories that exhibit the same behavioral patterns as those observed in the seed dataset, while the lower one contains those that are dissimilar. As our clustering algorithm iteratively extracts uni-behavior subsets from the multi-behavior dataset, theoretically, in the final clustering iteration, only one uni-behavior subset will remain dominant. Consequently, the resulting histogram may display either two distributions, with the lower one being very small, or just one distribution. Therefore, our approach to detecting the final cluster is to examine the number of trajectories within the lower probability distribution of the last histogram in the semi-supervised iterations (introduced in Section \ref{subsubsec:use-pu-filter}). If the number of trajectories is less than a specified count threshold, the corresponding  cluster is determined to be the final one and the clustering process terminates. A detailed example illustrating the determination of the last cluster is provided in Appendix \ref{append:subsec-end-of-clu}.

\begin{table*}[]
\centering
\caption{Comparison of ARI between the results of our approach and baselines, along with robustness testing results for our approach.}
\resizebox{\textwidth}{!}{
\begin{tabular}{l||ccccc||cc}
\multirow{2}{*}{} & \multicolumn{5}{c||}{Baseline Comparison}                                    & \multicolumn{2}{c}{Robustness test (\textbf{Ours})} \\ \cline{2-8}
                  & KM + $\bm{a}$ & KM + $(\bm{s,a,s'},r)$ & KM + TAAT & DBSCAN + TAAT & \textbf{Ours}  & Imbalanced     & Noise    \\ \hline\hline
Loco- Ant        & 0.086 & 0.089 & 0.866 & 0.223 & 0.971          & 0.963 & 0.971 \\
Loco-Halfcheetah & 0.102 & 0.104 & 0.818 & 0.143 & 0.992          & 0.985 & 0.972 \\
Loco-Hopper      & 0.036 & 0.034 & 0.991 & 0.502 & 0.999          & 1.000 & 0.993 \\
Loco-Walker2d    & 0.067 & 0.035 & 0.964 & 0.389 & 0.993          & 0.989 & 0.965 \\
Loco-Humanoid    & 0.379 & 0.298 & 0.991 & 0.286 & 1.000          & 0.976 & 0.982 \\ \hline
Loco-Average     & 0.134 & 0.112 & 0.926 & 0.309 & \textbf{0.991} & 0.983 & 0.977 \\ \hline\hline
Hand-Hammer      & 0.298 & 0.215 & 0.656 & 0.001 & 0.985          & 0.988 & 0.982 \\
Hand-Door        & 0.423 & 0.475 & 0.913 & 0.019 & 0.992          & 0.989 & 0.993 \\
Hand-Pen         & 0.274 & 0.028 & 0.519 & 0.035 & 0.984          & 0.937 & 0.969 \\ \hline
Hand-Average     & 0.332 & 0.239 & 0.696 & 0.018 & \textbf{0.987} & 0.971 & 0.981 \\ \hline\hline
Trifinger-Push   & 0.213 & 0.017 & 0.508 & 0.136 & 0.996          & 0.985 & 0.978 \\
Trifinger-Lift   & 0.117 & 0.007 & 0.301 & 0.426 & 0.961          & 0.921 & 0.941 \\ \hline\hline
Trifinger-Average & 0.165  & 0.012    & 0.405        & 0.281  & \textbf{0.979} & 0.953               & 0.960         \\ \hline
Average          & 0.200 & 0.130 & 0.753 & 0.216 & \textbf{0.987} & 0.973 & 0.975
\end{tabular}
}
\label{table:cluster-results}
\end{table*}

\section{Experiments and results}

\subsection{Baseline comparison to our approach} \label{subsec:baselines}
\vspace{-1.0mm}
\begin{itemize}[leftmargin=*]
    \item K-means + $\bm{a}$: K-means is applied to the raw action set of the multi-behaviour dataset for clustering. The exact correct cluster number is predefined for K-means, giving K-means an unfair advantage compared to our approach.
    \vspace{-1mm}
    \item K-means + $(\bm{s,a,s'},r)$: As in previous work \cite{clue, reducing, cer}, K-means is applied to the transition data. Here, $\bm{s'}$ represents the next state after taking action $\bm{a}$ at state $\bm{s}$, and the agent acquires the reward $r$. These vectors, $\bm{s'}$, $\bm{s}$, $\bm{a}$ and scalar $r$ are concatenated to form vectors for K-means clustering and the cluster number is also predefined.
    \vspace{-1mm}
    \item K-means + TAAT: K-means is applied to the long-horizon TAAT (obtained using Equation \ref{eq: action-rep}) of the trajectories. The cluster number is also predefined.
    \vspace{-1mm}
    \item DBSCAN + TAAT: The density-based clustering algorithm DBSCAN \cite{dbscan} is applied to the TAAT of the trajectories. DBSCAN has the advantage of not requiring a predefined cluster number, unlike K-means. However, because DBSCAN is computationally inefficient, it is not applied to the raw dataset with a large number of samples as it would take an extremely long time. The crucial parameters of DBSCAN, the neighborhood radius ($\epsilon$) and the minimum number of points ($\phi$), are optimized through a grid search. This involves testing $\epsilon$ values ranging from $0.1$ to $2$, evenly spaced with $20$ different values within this range, as well as exploring $\phi$ values from $1$ to $20$, resulting in a total of $400$ grid search combinations.
\end{itemize}


\subsection{Robustness test of our approach}
We test the robustness of our approach on datasets from more realistic settings to mimic some real-world applications, including:
\vspace{-3.0mm}
\begin{itemize}[leftmargin=*]
 \item Imbalanced datasets: For the multi-behavior datasets which is the union of $6$ uni-behavior subsets, we adjust the ratio of uni-behavior subset cardinalities to 5:5:3:3:1:1; for those with $2$ uni-behavior subsets, we adjust the ratio to 5:1. Previously, they were all in equal proportions, as introduced in Section \ref{subsec:task-and-dsets}.
 \vspace{-2mm}
 \item Dataset with noise: Noise is introduced to the state and action data in the raw multi-behaviour dataset. In our testing, this noise includes two types, each accounting for 50\%: one is uniform noise with a random range of 5-20\% of the entire state/action space (minimum values to maximum values), and the other is Gaussian noise with a mean of 0 and a standard deviation of 5-20\% of the entire state/action space.
\end{itemize}

\subsection{Results}
The Adjusted Rand Index (ARI) \cite{adjust-rand-score} is utilized as the evaluation metric in our work. A higher ARI value implies better clustering results. The results are reported in Table \ref{table:cluster-results}.

When using K-means for clustering directly on the raw data, including both action and transition data, the results are poor, even though these results were obtained when we specified the correct number of clusters for K-means, giving it an unfair advantage. Furthermore, we found that employing transition data does not enhance the effectiveness of K-means compared to using only action data in our settings.

When performing clustering on the TAAT data, K-means demonstrates satisfactory results, especially on the dataset of the locomotion tasks, where it even achieved a high average ARI of $0.926$. Although DBSCAN is more flexible than K-means, since it does not require a predefined cluster number, it does not perform optimally, even if utilizing our TAAT and the optimized parameters. This suggests that DBSCAN may not be suitable for handling the policy dataset, possibly due to non-uniform data density of clusters.

Our clustering approach demonstrates excellent performance across all datasets with an average ARI of $0.987$. Furthermore, it exhibits robustness, as it is just slightly affected by imbalanced and noisy datasets.

\section{Advantages of using clustering for offline policy learning}
\subsection{Learning policy from the clustered subsets}
We use the clustering datasets for offline policy learning. We separately train policies on data from each individual cluster obtained using our clustering method, then select the best-performing policy and compare it with the policy trained on the multi-behavior datasets. We report the results in Appendix \ref{appendix:advantages-using-clustering-for-policy}; as policy learning is not the focus of this paper, we placed the policy training results there to save page space of the main text. It can be seen that best policies trained on clustered subsets can outperform those trained on multi-behavior datasets, even though the multi-behavior datasets have six times more data than the clustered subsets.

We want to emphasize the advantages that our method brings to DIL, i.e., behavioral cloning \cite{bc}. Most DIL algorithms possesses the characteristics of supervised learning, which involves learning a regression function to establish a mapping from states to actions. However, if the dataset contains multiple behavior patterns, it is likely to have several different action (distributions) outputs for the same state. This scenario can lead to ambiguity during the regression process, thereby impairing the effectiveness of policy learning. Our method addresses this issue by clustering the data based on behavior patterns. Theoretically, the actions, when conditioned on states, exhibit a unimodal distribution across the clustered datasets, which aids in avoiding the aforementioned ambiguity risk in DIL. Most importantly, our clustering algorithm shares the \textbf{reward-free} attribute common to most DIL algorithms, enabling policy learning in datasets entirely devoid of rewards and thus avoiding the need for a complex reward design process.


\subsection{Policy ensemble}
Solely using datasets that train strong policies, while neglecting others, leads to the underutilization of data resources; this is sometimes unacceptable, particularly when the dataset is of small size. One way to optimize data utilization is by integrating clustered datasets through ensemble policy learning  \cite{ensemble-learning-survey, ensemble-rl-survey}, which has been employed in several prior policy learning projects and has shown effective outcomes in both online \cite{ensemble-online1, emsemble-online2, emsemble-online3} and offline learning contexts \cite{ensemble-offline-rl-edac, ensemble-offline-rl-pebl, ensemble-offline-rl-q-ensemble}. Specifically, research can be directed towards several areas, including: ensembles of reward functions; ensembles of value functions; ensembles of loss functions; and ensembles of decisions involving techniques such as weighted voting \cite{ensemble-vote2}, weighted aggregation \cite{ensemble-agg1}, model stacking \cite{ensemble-stack}, and model selection \cite{ensemble-optimal}. 

\section{Conclusion}
This paper introduces a simple deep clustering method capable of grouping multi-behavior datasets into uni-behavior subsets to benefit downstream policy learning. Our clustering approach demonstrates high effectiveness and robustness. It is important to highlight that our approach is not limited to the single-task datasets presented in this paper; it has the potential to be adapted for other prominent research challenges, including multi-task learning \cite{multi-task} and multi-agent learning \cite{multi-agent}.

Nonetheless, there are still some limitations to our method. Firstly, due to the necessity to training multiple deep learning-based neural networks, the computational cost of clustering operations is high. Secondly, when terminal points of the trajectory are not specified, our method is limited in finding effective clustering feature, which could affect the subsequent clustering. In our future work, we aim to extend our method to encompass more practical datasets and applications. For instance, this could involve incorporating human demonstration datasets or developing policy learning algorithms to complement our clustering approach.

\bibliography{manual}
\bibliographystyle{icml2023}

\newpage
\appendix
\onecolumn
\section{Uni-behavior dataset versus multi-behavior dataset for offline policy learning} \label{append:multi-vs-uni-results}
This section aims to support the motivation of this paper proposed in the Section \ref{sec:intro}, which suggests that offline policy learning is more likely to acquire satisfactory policies from uni-behavior datasets than from multi-behavior datasets. Here, we collected $3$ separate uni-behavior datasets and subsequently merged them into a multi-behavior dataset. These datasets were collected from $4$ different locomotion control tasks in the D4RL benchmark \cite{d4rl}, including \textit{Halfcheetah}, \textit{Ant}, \textit{Hopper}, and \textit{Walker2d}. Three uni-behavior datasets were designed to generally have a similar skill level to ensure that the multi-behavior dataset does not endow an unfair advantage or disadvantage in terms of the skill level of the learned policy compared to the uni-behavior datasets. More details about the compositions of the datasets refer to Appendix \ref{append:dset-compo}.

\begin{figure}[!htbp]
    \centering
        \subfigure[HalfCheetah]{\includegraphics[width=0.95\textwidth]{ 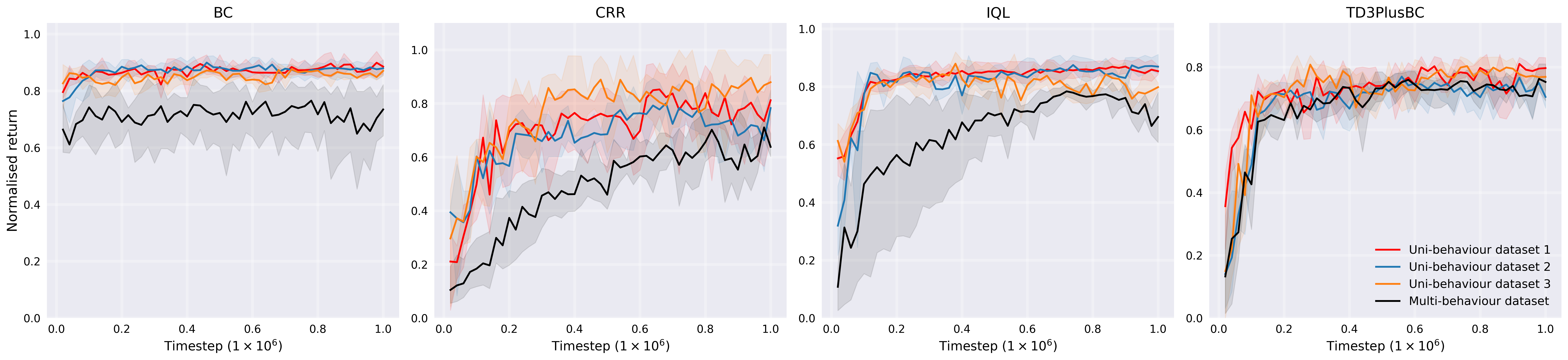}}
        \vskip -0.01mm
        \subfigure[Ant]{\includegraphics[width=0.95\textwidth]{ 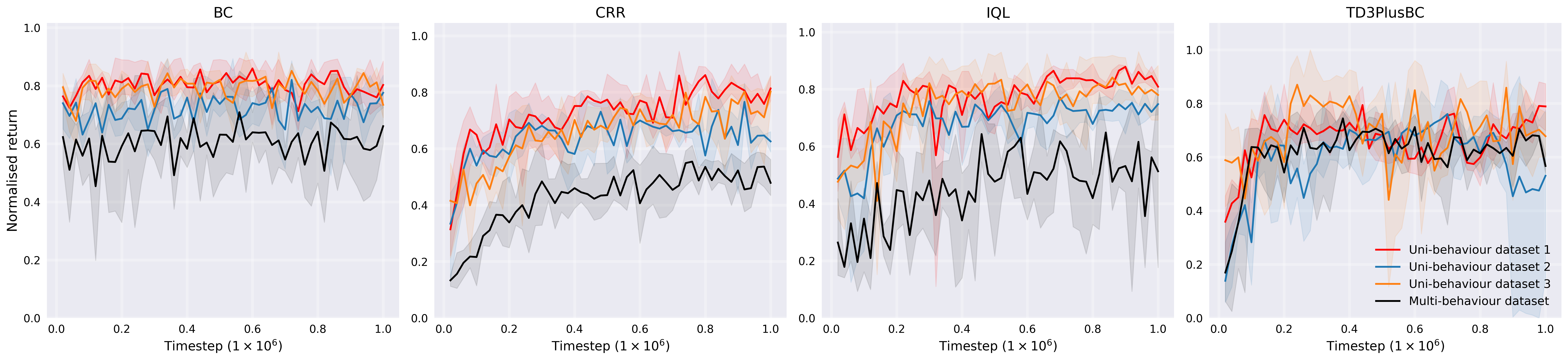}}
        \vskip -0.01mm
        \subfigure[Hopper]{\includegraphics[width=0.95\textwidth]{ 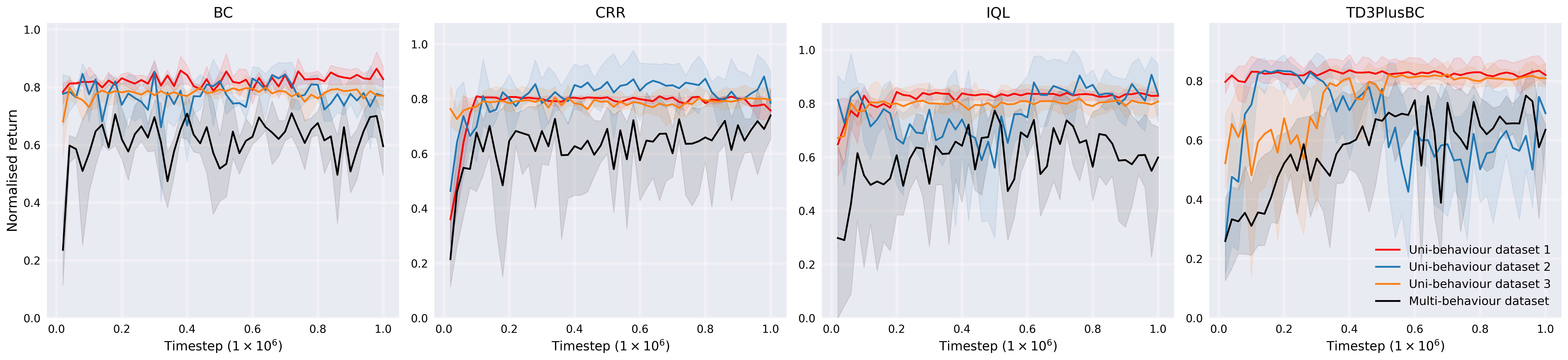}}
        \vskip -0.01mm
        \subfigure[Walker2D]{\includegraphics[width=0.95\textwidth]{ 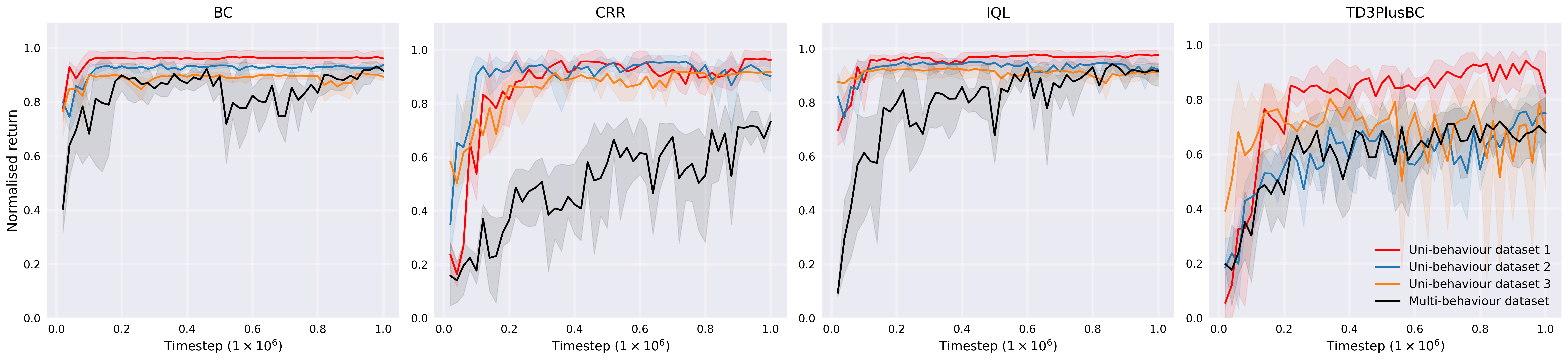}}
    \caption{Illustration of the performance of agents trained using multi-behavior and uni-behavior datasets with different algorithms on various tasks. Each data point on the plots represents evaluation results from 5 episodes, and the scores are normalized using: $score_{norm} = ({score-score_{min}})/({score_{max}-score_{min}})$.}
    \label{fig:single-vs-multi}  
\end{figure}

Then, we utilize these datasets for offline policy learning, employing $4$ different types of algorithms:
\begin{itemize}
    \item Behavioral Cloning (BC) \cite{bc}: It is a offline deep imitation learning (DIL) algorithm based on supervised learning. Its goal is to learn a policy to mimic behaviors in a dataset by mapping states to actions. Its training objective can be represented as: $\min_{\pi} \mathbb{E}_{(s_{i},a_{i}) \sim \mathcal{D}} \left[(a_{i} - \pi(s_{i}) )^2\right]$.
    \item Critic Regularized Regression (CRR) \cite{crr}: This is an offline deep reinforcement learning (DRL) algorithm that fundamentally operates as a weight-based BC. It begins by training a critic, which is then utilized to estimate the advantage of the training samples in the dataset. These estimates are then employed as weights for BC.
    \item Implicit Q-Learning (IQL) \cite{iql}: It is an offline DRL algorithm that employs an indirect and implicit approach to estimate the value of actions, helping to mitigate the risk of overestimating less common actions in the training dataset. This is essential for addressing the common out-of-distribution (OOD) problem in offline DRL.
    \item TD3PlusBC \cite{td3bc}: It is an offline DRL algorithm that modifies the online DRL algorithm TD3 \cite{td3} by incorporating BC as a explicit regularization term in the action output function. This change ensures the learned actor function aligns more closely with the dataset's action distribution, hence addressing OOD issues in offline DRL.
\end{itemize}

The results are presented in Figure \ref{fig:single-vs-multi}, where it is evident that agents trained on multi-behavior datasets generally perform worse than those trained on uni-behavior datasets in terms of learning speed, stability and performance. It should be noted that the exploration diversity here, which encompasses the covered state, action, and reward space of the multi-behavior datasets, appears to be larger than that of the uni-behavior datasets. This exploration diversity is crucial for offline policy learning \cite{offline-rl-noise}. Additionally, the size of the multi-behavior datasets is also larger than that of the uni-behavior datasets. Despite these advantages, policies trained on multi-behavior datasets still tend to perform worse.

\section{Proofs of observations, assumptions, and equations in the main text}
\subsection{Proof of Observation \ref{observation: 1}}  \label{append:subsec:obs1}
This section aims to validate the Observation \ref{observation: 1}. We compute the mean of the lower percentile of pairwise Euclidean distances between actions of the same behaviour policy; this can be expressed as: $ \delta_{same} = \mathbb{E}_{a_{1}, a_{2} \sim \mathcal{D}^{\mathcal{A}}_{\pi_p}} [\mathcal{P}(d(a_{1}, a_{2}))]$. Similarly, we compute the mean pairwise Euclidean distances between actions from different behaviour policies and then filter them by percentile; this can be expressed as: $ \delta_{diff} = \mathbb{E}_{a_{1} \sim \mathcal{D}^{\mathcal{A}}_{\pi_p}, a_{2} \sim \mathcal{D}^{\mathcal{A}}_{\pi_q}} [\mathcal{P}(d(a_{1}, a_{2}))], \pi_{p} \ne \pi_{q}$ (notations are introduced in Section \ref{subsec:analy-action-data}). In the equations above, $\mathcal{P}$ represents the percentile filtering operator, and $d$ denotes the calculation of Euclidean distance. The experimental data used here is from Section \ref{subsec:task-and-dsets}. Specifically, we sampled 5,000 action data points from each uni-behavior dataset to conduct the above calculations.

\begin{figure}[!htbp]
    \centering
        \subfigure[Hand]{\includegraphics[width=0.3\textwidth]{ 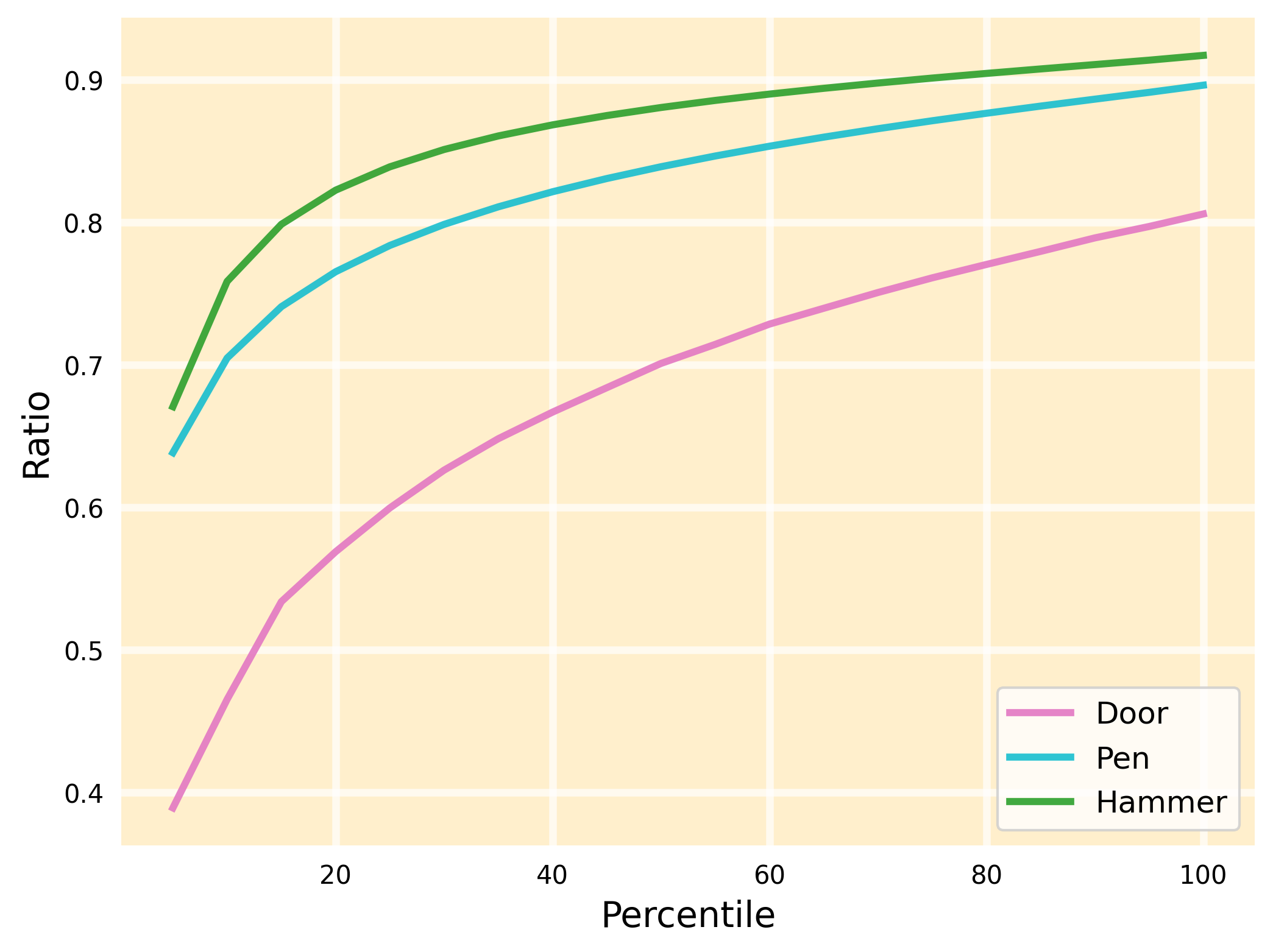}}
        \subfigure[Locomotion]{\includegraphics[width=0.3\textwidth]{ 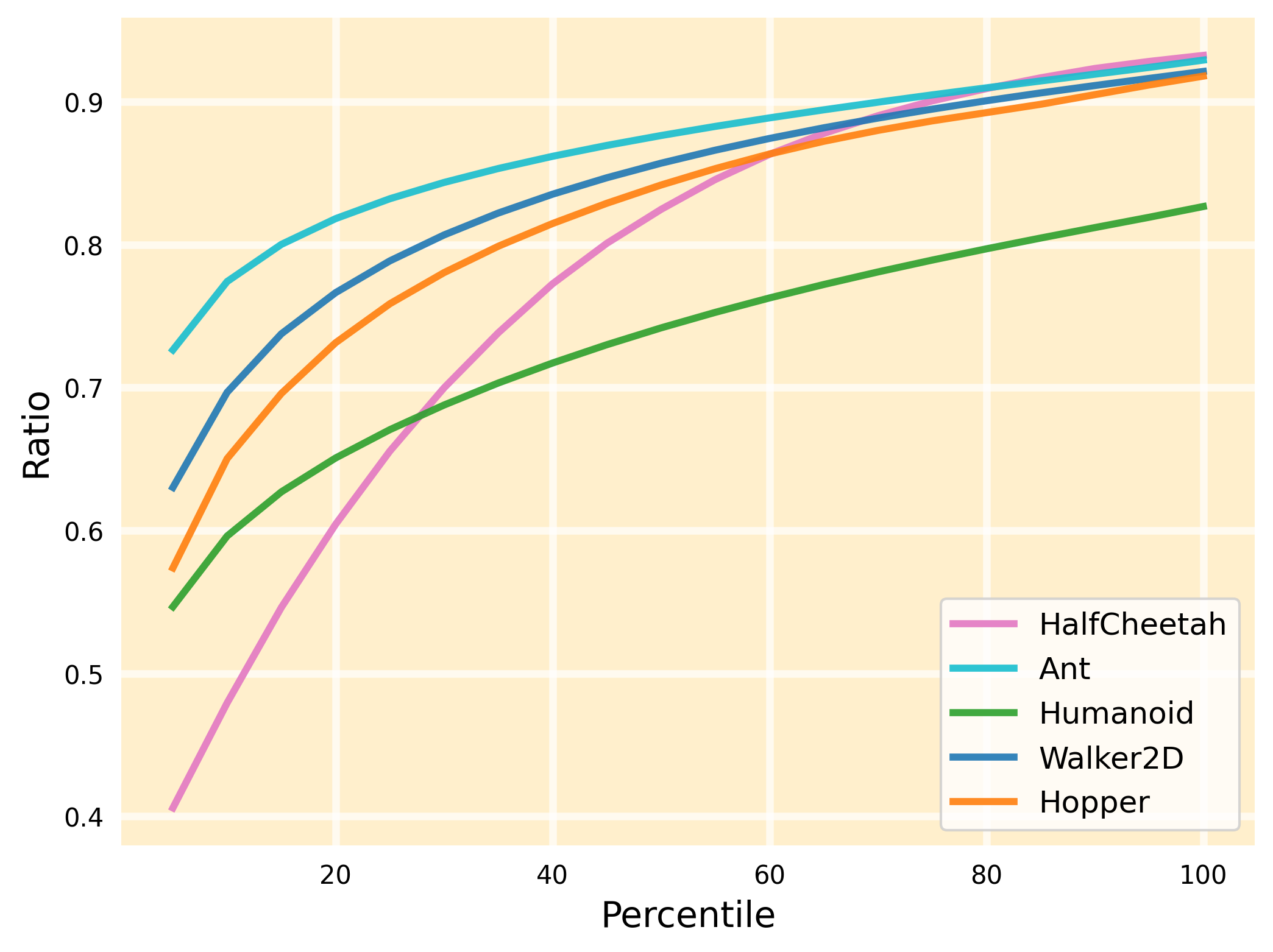}}
        \subfigure[TriFinger]{\includegraphics[width=0.304\textwidth]{ 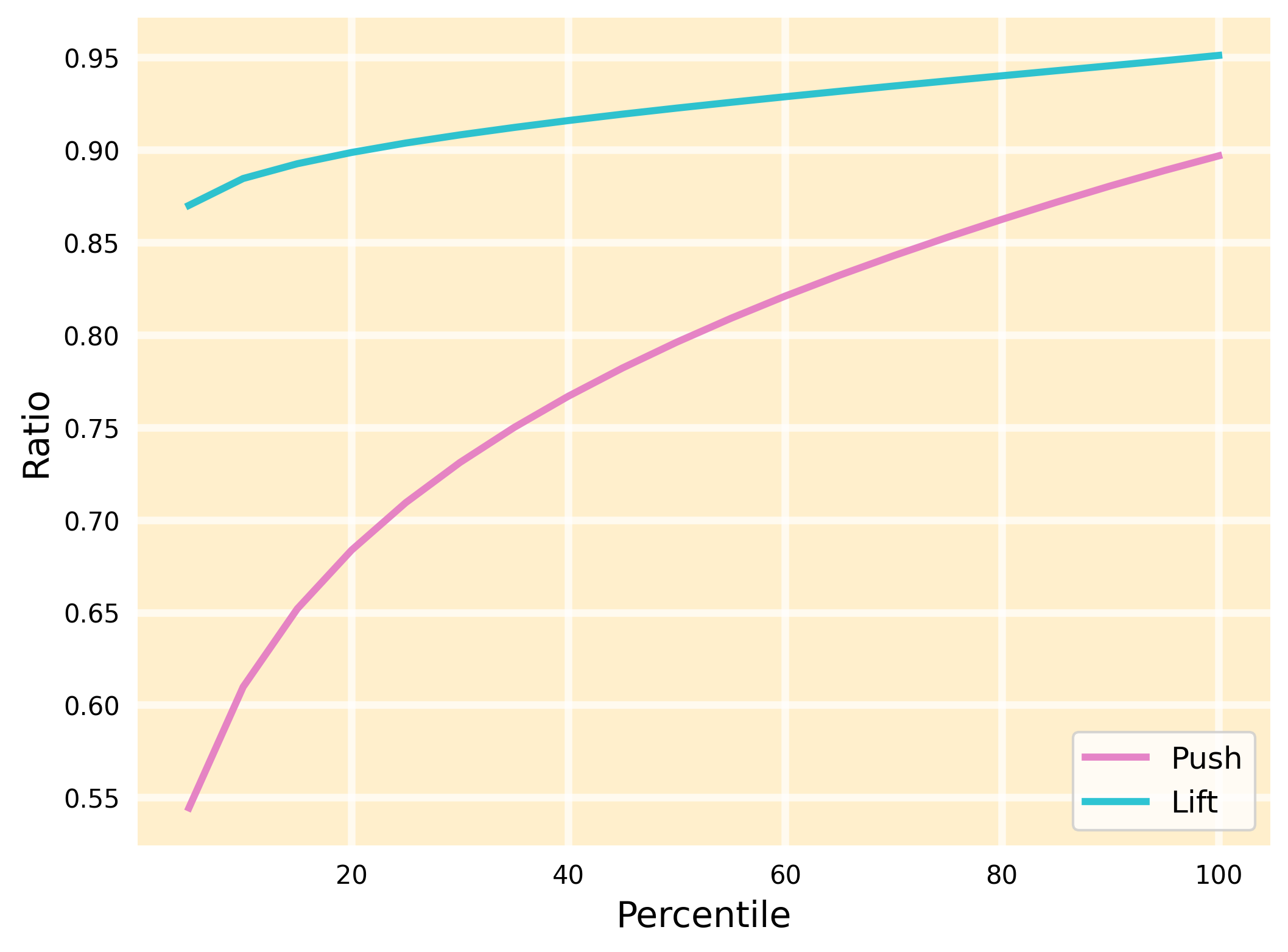}}
    \caption{The ratio of \(\delta_{\text{same}}/\delta_{\text{diff}}\) across a range of percentile values in ten distinct datasets spanning three different benchmark suites.}
    \label{fig:obs1-percentil}  
\end{figure}

\begin{figure}[!htbp]
\begin{center}
        \subfigure[Hand-Hammer]{
                \includegraphics[width=0.24\textwidth]{ 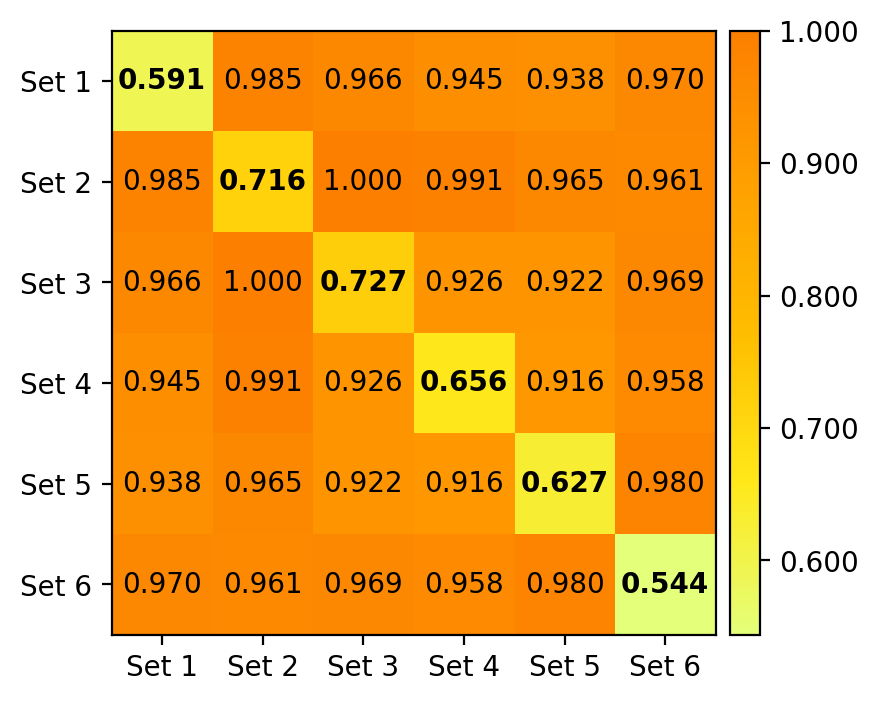}
                }
        \hskip -0.1in
        \subfigure[Hand-Door]{
                \includegraphics[width=0.24\textwidth]{ 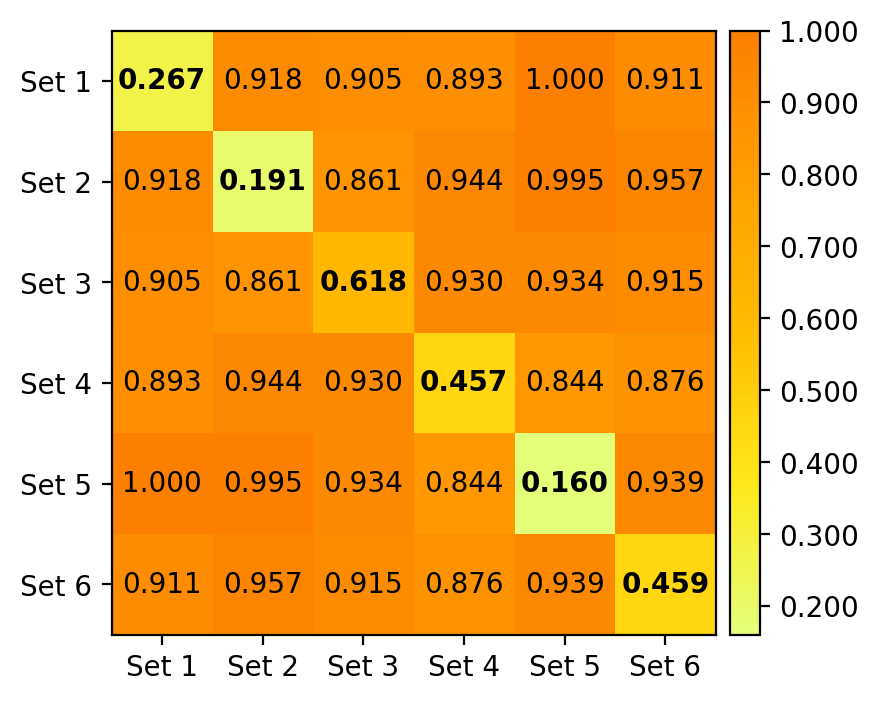}
                }
        \hskip -0.1in
        \subfigure[Hand-Pen]{
                \includegraphics[width=0.24\textwidth]{ 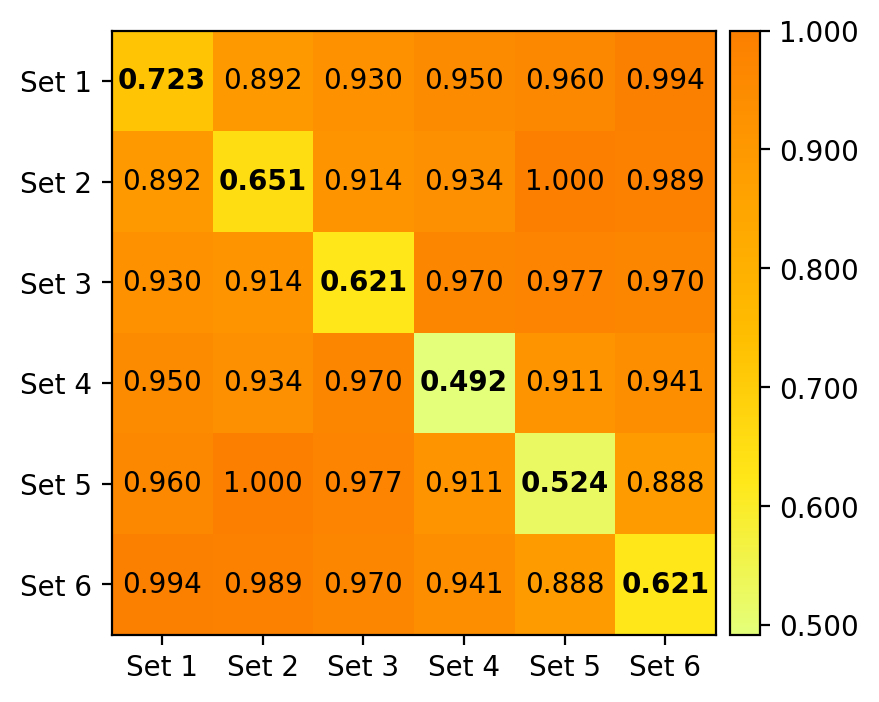}
                }
        \hskip -0.1in
        \subfigure[Loco-HalfCheetah]{
                \includegraphics[width=0.24\textwidth]{ 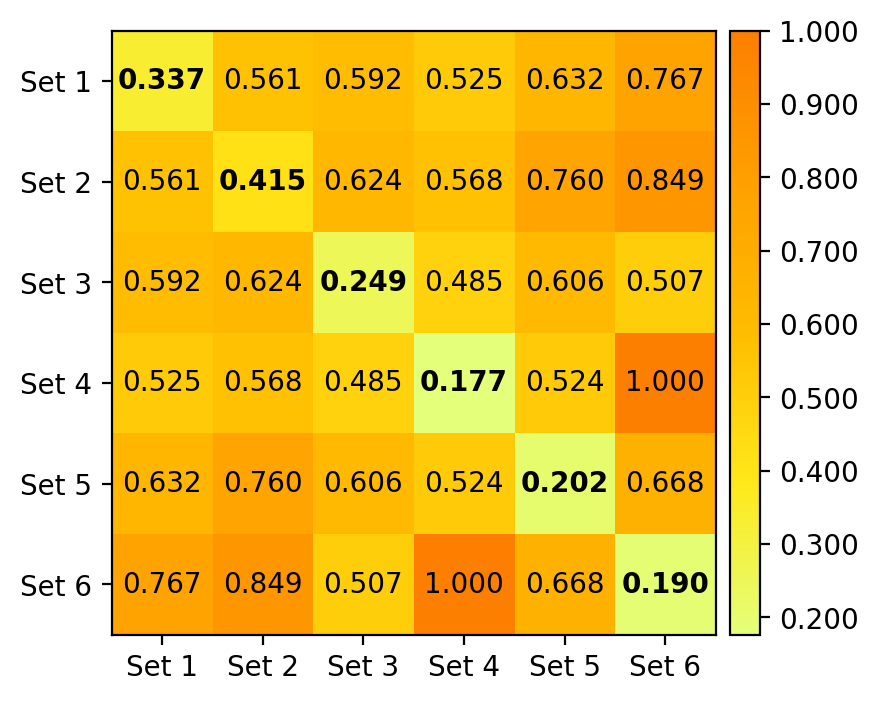}
                }
                
    \vskip -0.1in
        \subfigure[Loco-Walker2D]{
                \includegraphics[width=0.24\textwidth]{ 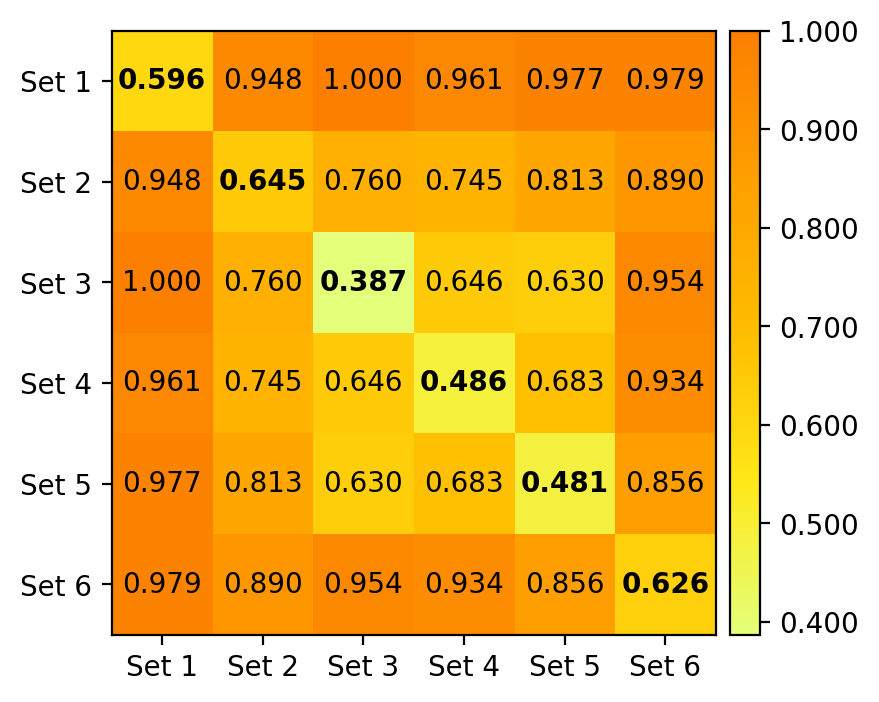}
                }
        \hskip -0.1in
        \subfigure[Loco-Hopper]{
                \includegraphics[width=0.24\textwidth]{ 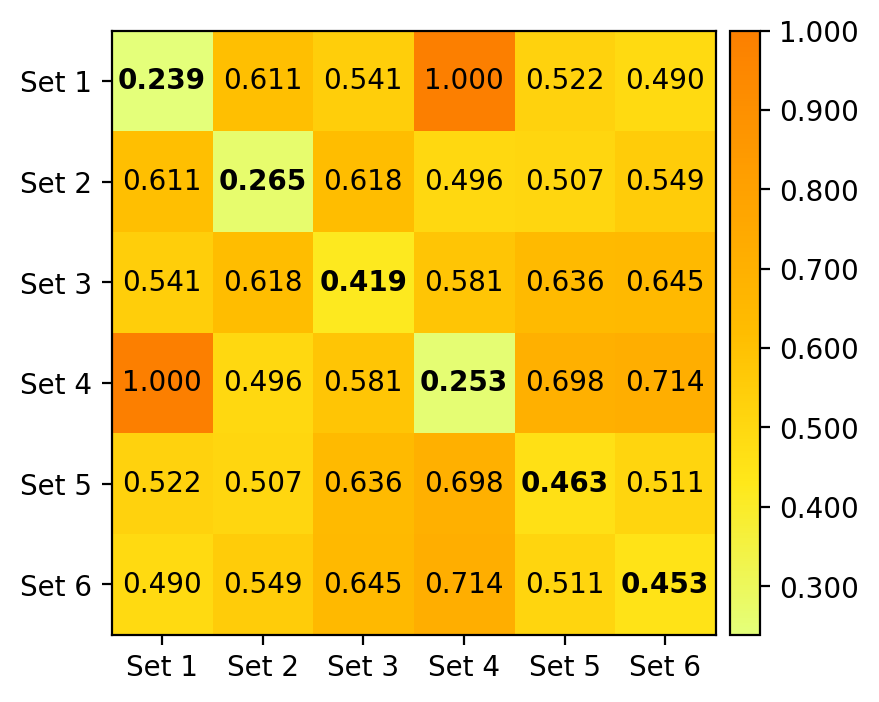}
                }
        \hskip -0.1in
        \subfigure[Loco-Ant]{
                \includegraphics[width=0.24\textwidth]{ 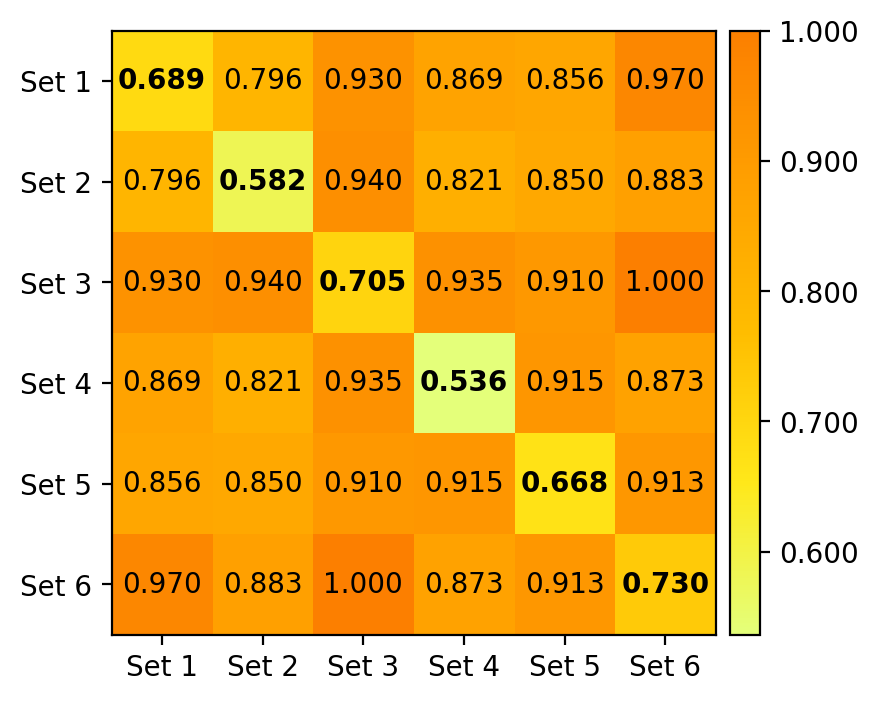}
                }
        \hskip -0.1in
        \subfigure[Loco-Humanoid]{
                \includegraphics[width=0.24\textwidth]{ 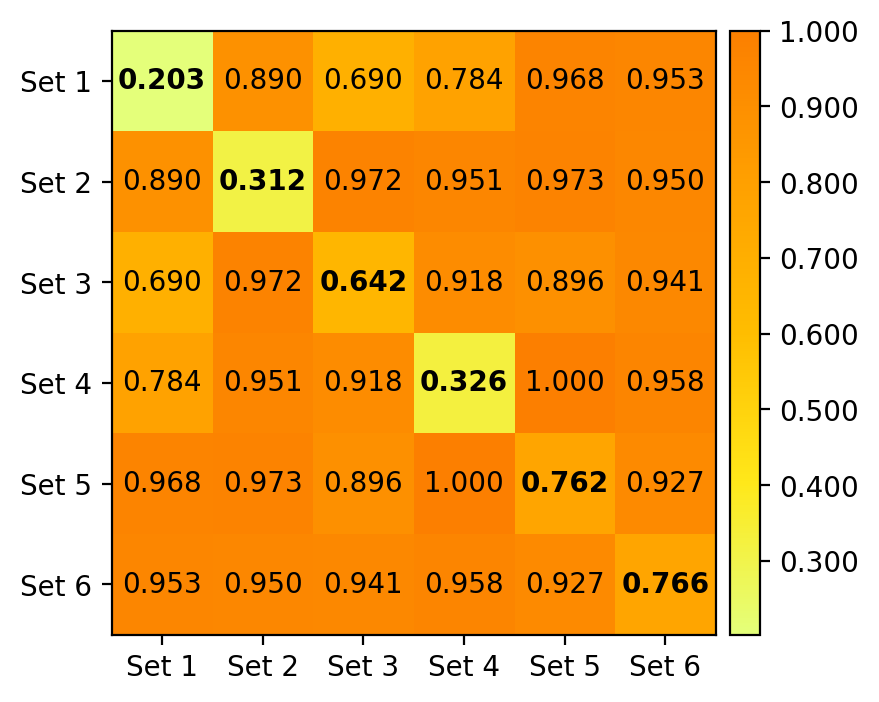}
                }
                
    \vskip -0.1in
        \subfigure[Trifinger-Push]{
                \includegraphics[width=0.24\textwidth]{ 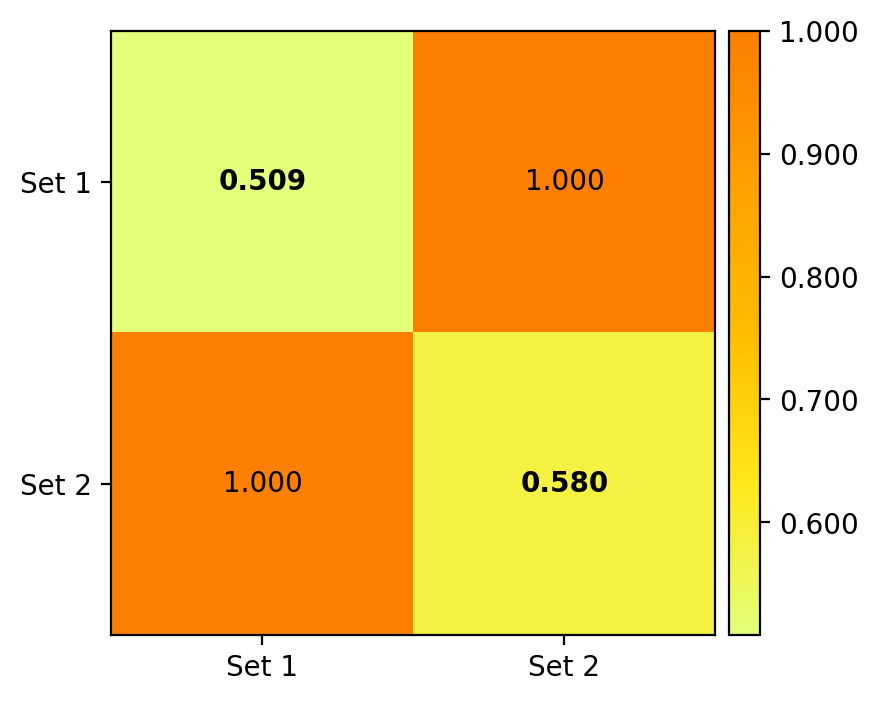}
                }
        \hskip -0.1in
        \subfigure[Trifinger-Lift]{
                \includegraphics[width=0.24\textwidth]{ 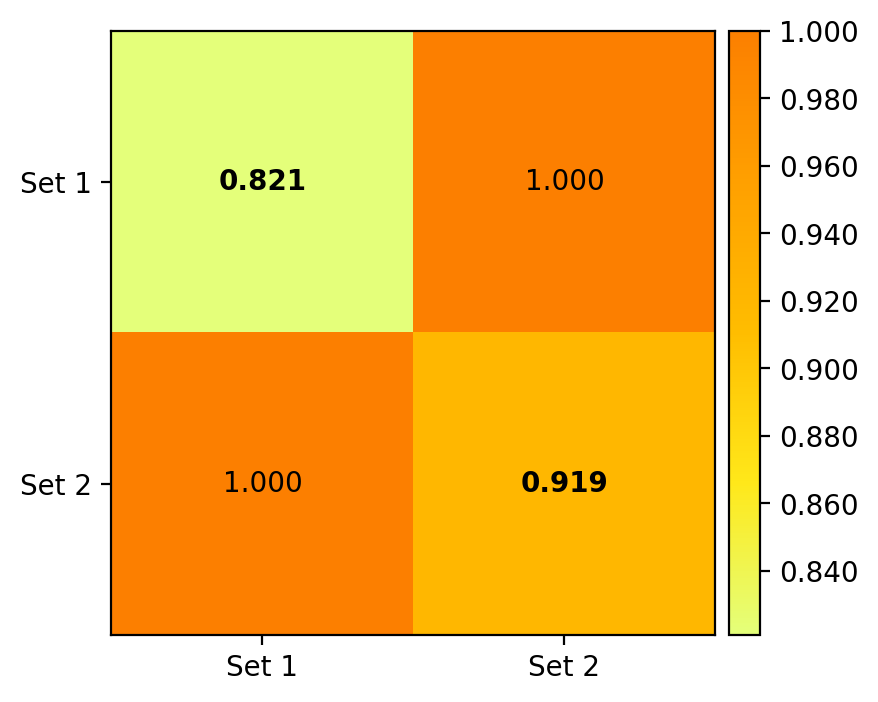}
                }
    \caption{Mean pairwise Euclidean distances filtered using a percentile threshold of 5\%. These distances are computed between actions within the same uni-behavior dataset (represented by the numbers on the diagonal) and between actions from different uni-behavior datasets (represented by the numbers off the diagonal).}
    \label{fig:obs1}  
\end{center}  
\end{figure}

In Figure \ref{fig:obs1-percentil}, we present the ratio of $\delta_{same}/\delta_{diff}$ across a range of filter percentile values, from 5\% to 100\%. A percentile value of 100\% indicates that there is no percentile filtering applied, and all data is used. Lower $\delta_{same}/\delta_{diff}$ ratio values imply a more concentrated trend in the action samples selected under specific percentile thresholds from their respective policies. Conversely, higher ratio values suggest a tendency towards a more overlapping distributions of action samples from different policies. From the figure, we can see that the ratio approaches one when no percentile filtering is used. This indicates that the mean pairwise distance between actions from the same policy and actions from different policies approaches the same value, implying that actions from various policies are essentially uniformly mixed. In contrast, each policy's output actions have their respective concentrated areas, as the ratio tends to be smaller when filtering with lower percentile thresholds. Figure \ref{fig:obs1} presents detailed results for the 5\% percentile. It is observed that action data from the same policy tends to group closely together, potentially forming a high-density regions of the actions space. Of course, in practice, when filtering action pairs based on their pairwise distances, we do not know if the pair are both from the same uni-behavior distribution or the from different uni-behavior distributions in the action space. 


\subsection{Proof of Equation \ref{eq: action-rep}} \label{append:prove-eq}
\subsubsection{Mathematical analysis}
Equation \ref{eq: action-rep} can be intuitively interpreted as applying the Weak Law of Large Numbers (WLLN), that is:

{\textbf{Weak Law of Large Numbers (Chebyshev's Version)}} \textit{Let \(X_1, X_2, \ldots, X_n\) be a sequence of independent and identically distributed random variables, each with a finite expected mean \(E(X_i) = \mu\) and finite variance \(Var(X_i) = \sigma^2\). Define the sample mean of these variables as \(\overline{X}_n = \frac{1}{n}(X_1 + X_2 + \ldots + X_n)\). The Weak Law of Large Numbers (WLLN) posits that for any positive number \(\beta\), as the number of observations \(n\) increases, the probability that the sample mean deviates from the true mean by less than \(\beta\) tends towards $1$, formally: $\lim_{n \to \infty} P(|\overline{X}_n - \mu| < \beta) = 1$. }

In this context, the sample mean of variables refers to the \textit{temporal-averaged action trajectory} (TAAT). The true mean refers to the population mean of action vector components; the WLLN suggests that as the number of sampled actions increases, the former will converge toward the latter. In our setting, the samples come from the same trajectory for feasibility. The application of the WLLN here relies on the strong assumption that all actions in the trajectory are independent and identically distributed. However, this assumption is difficult to verify in practice, due to variations in the length and distribution of the action data across different scenarios (i.e., different tasks or environments). Hence, we carry out the following experiments for further discussion.

\begin{figure}[!htbp]
    \centering
        \subfigure[Hand-Door]{
                \includegraphics[width=0.3\textwidth]{ 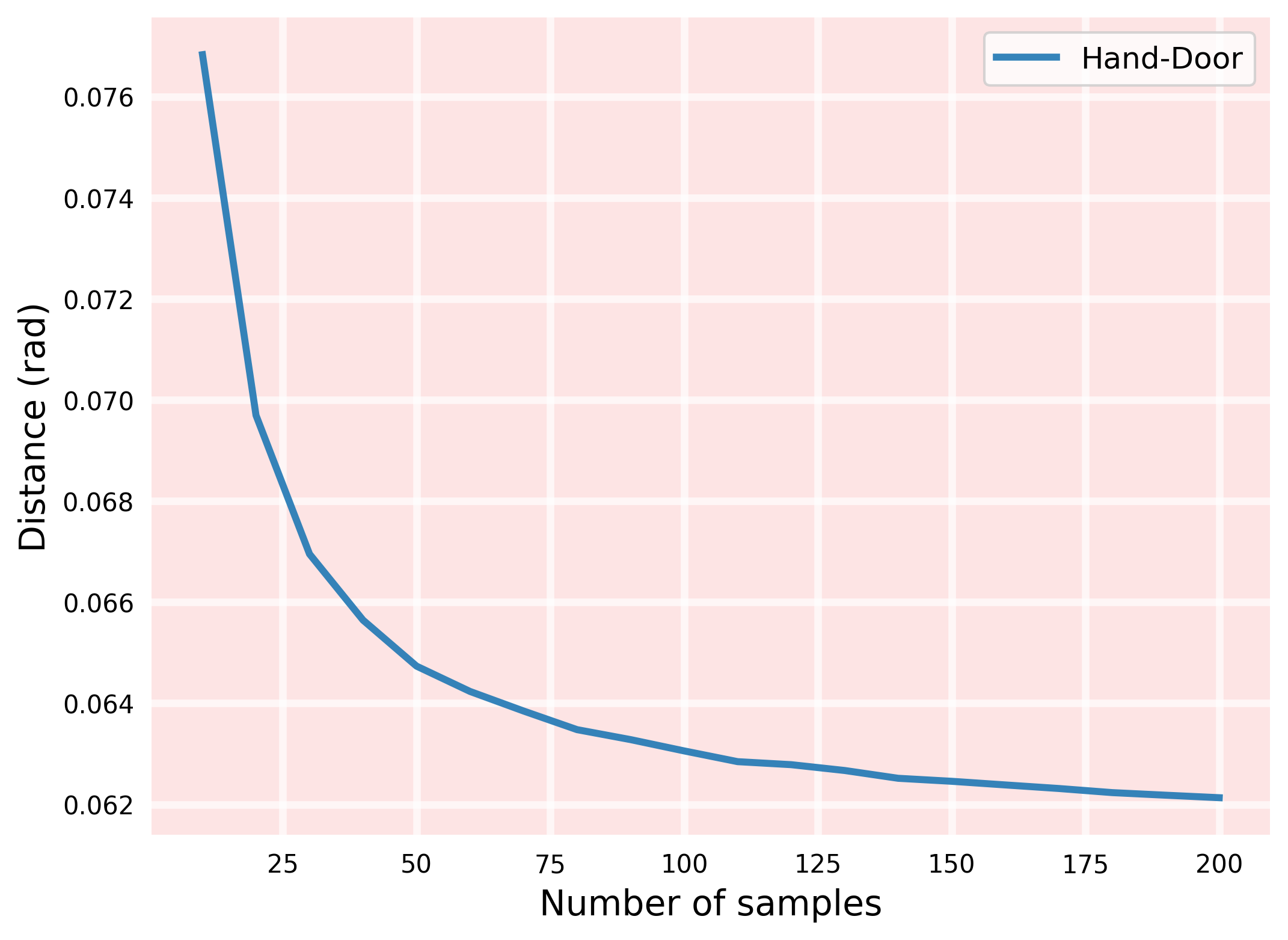}
                }
        \subfigure[Hand-Hammer]{
                \includegraphics[width=0.3\textwidth]{ 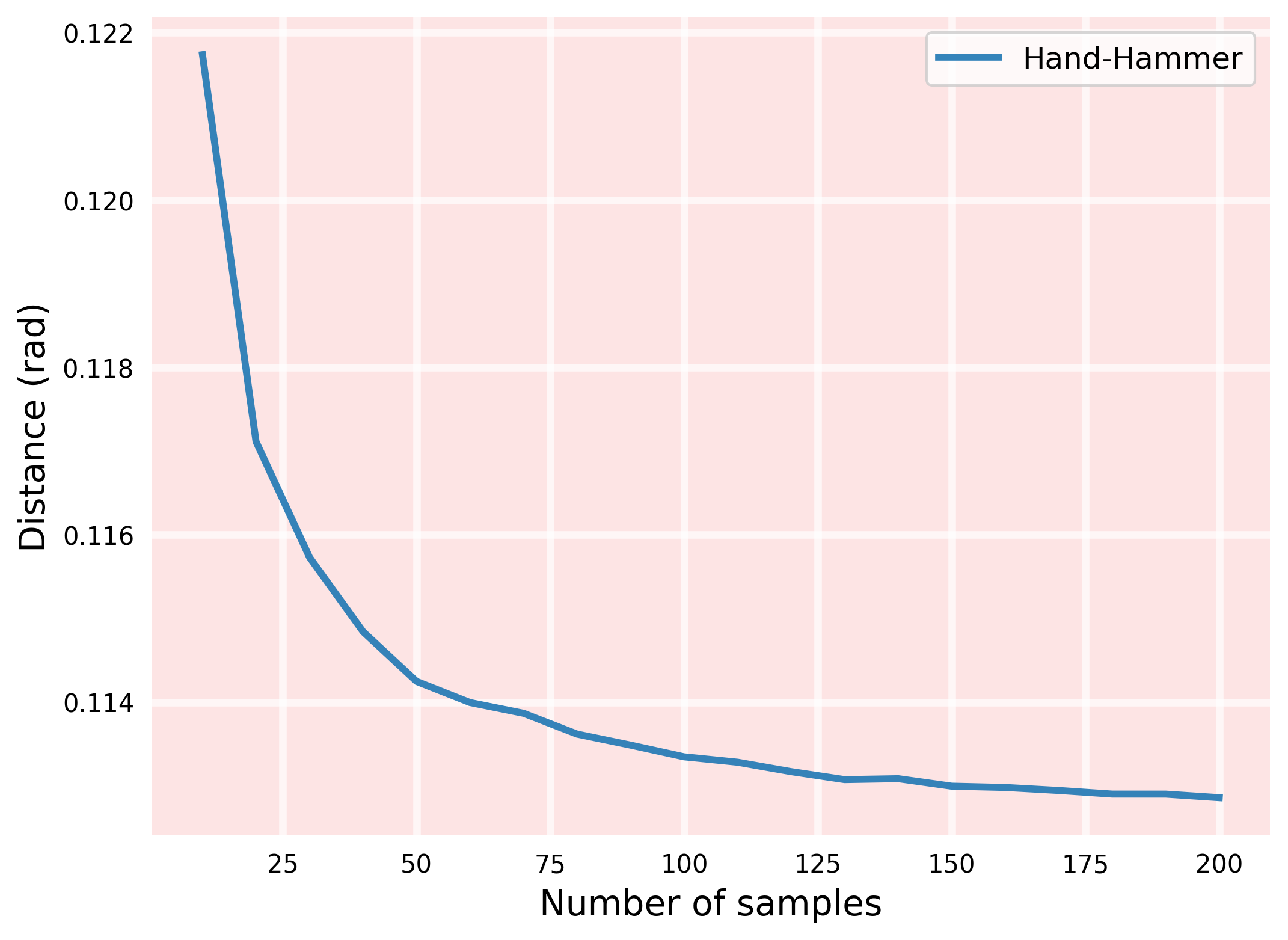}
                }
        \subfigure[Loco-HalfCheetah]{
                \includegraphics[width=0.3\textwidth]{ 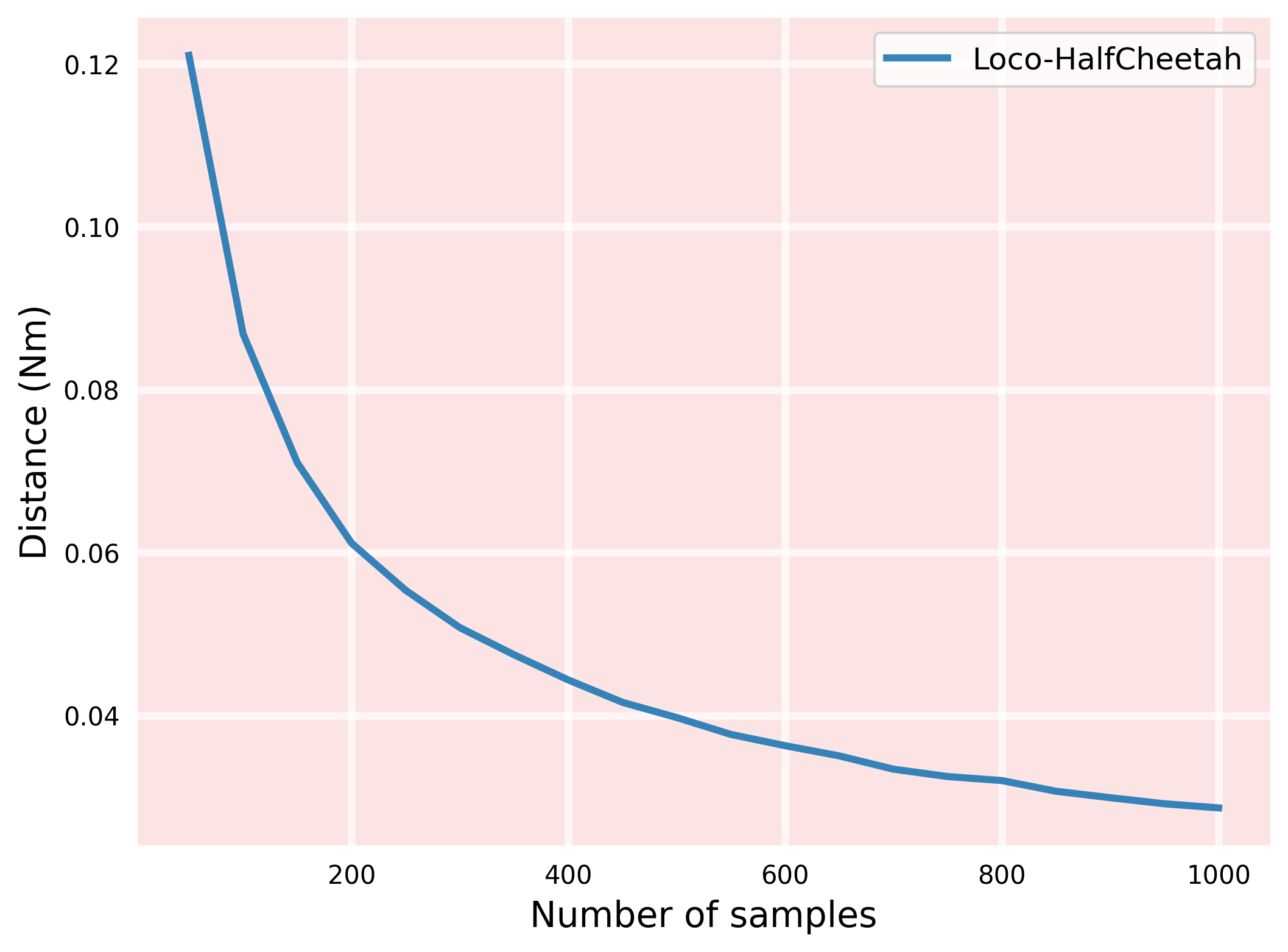}
                }
        \subfigure[TriFinger-Push]{
                \includegraphics[width=0.3\textwidth]{ 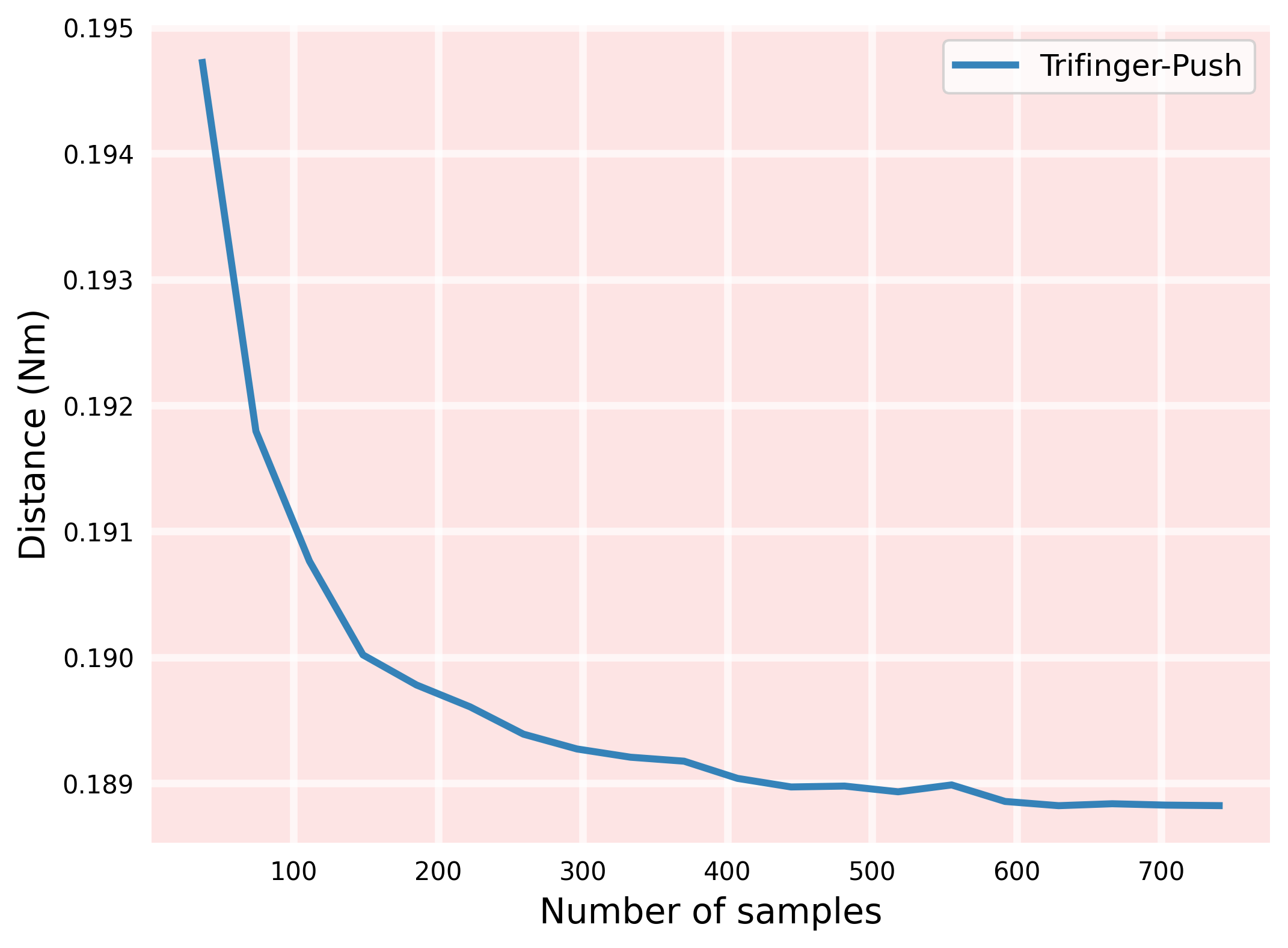}
                }
        \subfigure[TriFinger-Push]{
                \includegraphics[width=0.3\textwidth]{ 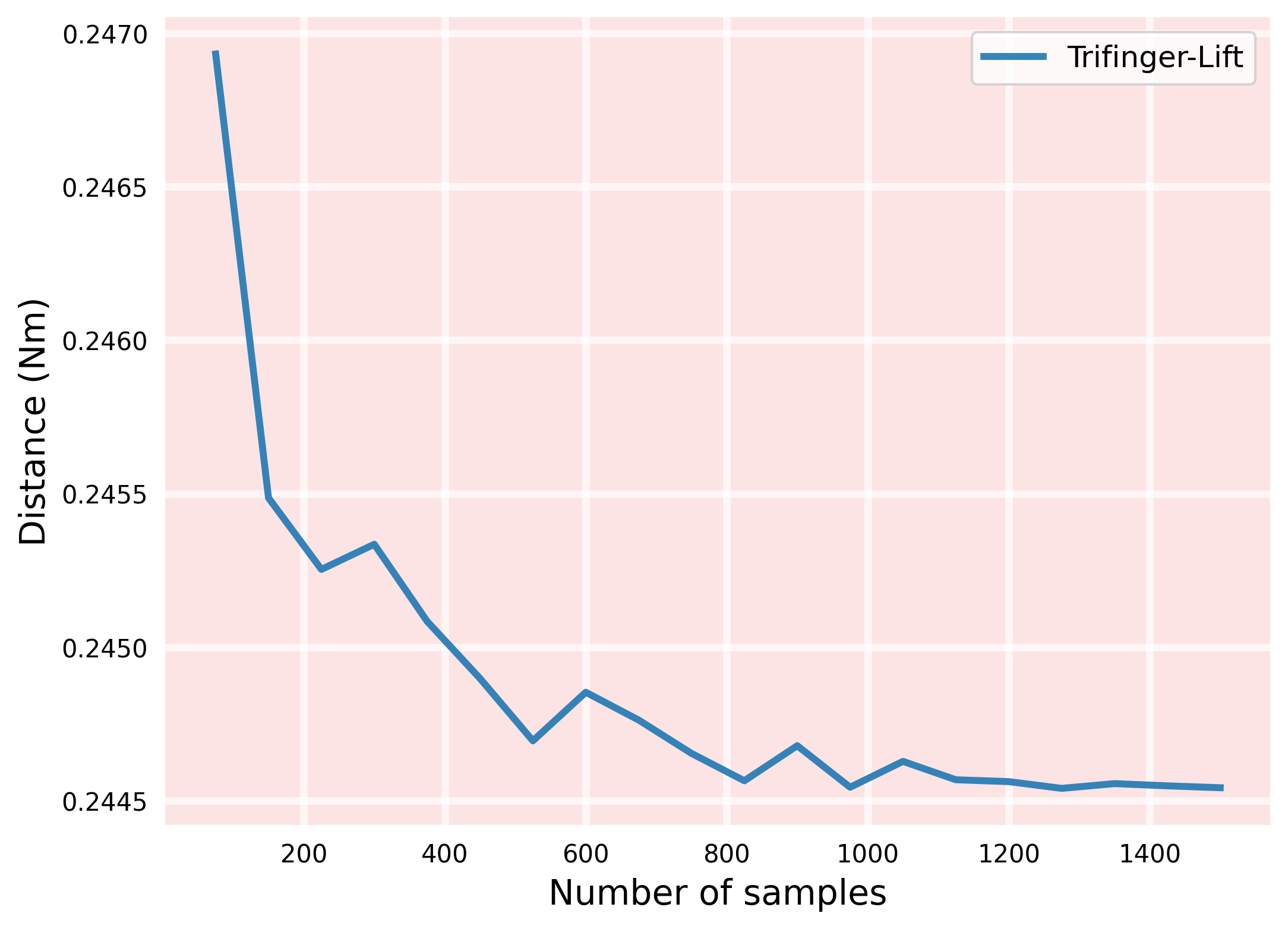}
                }
    \caption{Illustration of \(D_{\tau-\mu}\) (introduced in Appendix \ref{append:prove-eq}) over different trajectory durations for five different tasks. We chose these five tasks because they have fixed trajectory lengths according to the settings of the tasks, it facilitates obtaining trajectories of various lengths in our experiment.}
    \label{fig:eq9-wlln}
\end{figure}

In this experiment, our aim is to validate the convergence characteristics of the WLLN as applied to TAAT data; specifically, as the number of aggregated actions generating the TAAT increases, this TAAT will converge toward the true action cluster center. Firstly, we would determine the true action centre cluster. Directly determining the true cluster center is not feasible, so we approximate it by using the sample mean within the large-scale uni-behavior dataset, as referenced in Section \ref{subsec:task-and-dsets}. We denote this sample mean as \(\bm{\mu^{\mathcal{A}}}\). Then, we aggregate the trajectories using Equation \ref{eq: action-rep} to obtain the TAAT of each trajectory: $\{\bm{\overline{\tau_{1}^{\mathcal{A}}}}, \bm{\overline{\tau_{2}^{\mathcal{A}}}}, \ldots\}$, and calculate the mean Euclidean distance between the TAAT and the true clustering center using $D_{\tau-\mu} = \frac{1}{N} \sum_{i=1}^{N} d(\bm{\overline{\tau_{i}^{\mathcal{A}}}}, \bm{\mu^{\mathcal{A}}})$, where $N$ denotes the number of trajectories. We adjust the length of the trajectory to observe the variation in $D_{\tau-\mu}$, as depicted in Figure \ref{fig:eq9-wlln}. It can be observed that as the length of the trajectory increases, \(D_{\tau-\mu}\) gradually decreases. This suggests convergence characteristics as described in the WLLN. However, it is noticeable that \(D_{\tau-\mu}\) does not converge to zero. This may be due to the insufficient length of the trajectory and the fact that the actions are not independent and identically distributed, particularly when the actions within a trajectory are strongly correlated.

All in all, the above experiment proves that the TAAT obtained from Equation \ref{eq: action-rep} can converge towards the cluster center, and the longer the length, the closer the convergence.

\subsubsection{Quantitative findings}
A quantitative comparison of clustering trends in the action set before and after applying our TAAT is shown in Table \ref{table:validate-eq9}, illustrating the effectiveness of our TAAT method. We utilized three evaluation metrics: Silhouette score \cite{silhouettes}, Calinski-Harabasz index \cite{calinski-harabasz}, and Davies-Bouldin index \cite{davies-bouldin}. Note that higher Silhouette scores and Calinski-Harabasz index values, as well as lower Davies-Bouldin values, indicate more pronounced clustering trends among the data clusters, potentially leading to improved performance of clustering algorithms used later in the work to identify uni-behavior action clusters.
\begin{table*}[!htbp]
\centering
\caption{A quantitative comparison of clustering trends in the action set before and after employing our TAAT.}
\begin{tabular}{l||rr|rr|rr}
\multirow{2}{*}{} & \multicolumn{2}{c|}{Silhouette} & \multicolumn{2}{c|}{Calinski Harabasz} & \multicolumn{2}{c}{Davies Bouldin} \\ \cline{2-7} 
                    & TAAT           & No-TAAT & TAAT                & No-TAAT  & TAAT            & No-TAAT \\ \hline\hline
Loco-Ant          & 0.641          & -0.011  & 1802.917           & 103.899 & 0.602          & 4.899  \\
Loco-Halfcheetah  & 0.733          & -0.008  & 8061.308           & 59.783  & 0.538          & 14.586 \\
Loco-Hopper       & 0.730          & -0.066  & 51265.402          & 136.884 & 0.340          & 9.651  \\
Loco-Walker2d     & 0.657          & -0.021  & 10002.706          & 133.100 & 0.567          & 7.539  \\
Loco-Humanoid     & 0.656          & 0.102   & 8423.113           & 301.105 & 0.542          & 3.441  \\ \hline
Loco-Average      & \textbf{0.683} & -0.001  & \textbf{15911.089} & 146.954 & \textbf{0.518} & 8.023  \\ \hline\hline
Hand-Hammer       & 0.130          & 0.037   & 1188.860           & 139.607 & 2.461          & 4.100  \\
Hand-Door         & 0.399          & 0.136   & 4932.518           & 333.985 & 1.410          & 3.281  \\
Hand-Pen          & 0.122          & 0.049   & 2371.950           & 172.644 & 2.641          & 3.946  \\ \hline
Hand-Average      & \textbf{0.217} & 0.074   & \textbf{2831.109}  & 215.412 & \textbf{2.171} & 3.776  \\ \hline\hline
Trifinger-Push      & 0.214          & 0.093   & 953.260            & 407.924 & 1.883          & 3.489  \\
Trifinger-Lift      & 0.107          & 0.041   & 224.140            & 188.734 & 3.126          & 4.899  \\ \hline
Trifinger-Average & \textbf{0.161} & 0.067   & \textbf{588.700}   & 298.329 & \textbf{2.505} & 4.194  \\ \hline\hline
Average     & \textbf{0.439} & 0.035   & \textbf{8922.617}  & 197.767 & \textbf{1.411} & 5.983 
\end{tabular}
\label{table:validate-eq9}
\end{table*}

\subsection{Proof of Assumption \ref{assumption: 2}} \label{append:prove-assum2}
\begin{figure*}[htbp]
    \centering
        \subfigure[Hand]{\includegraphics[width=0.3\textwidth]{ 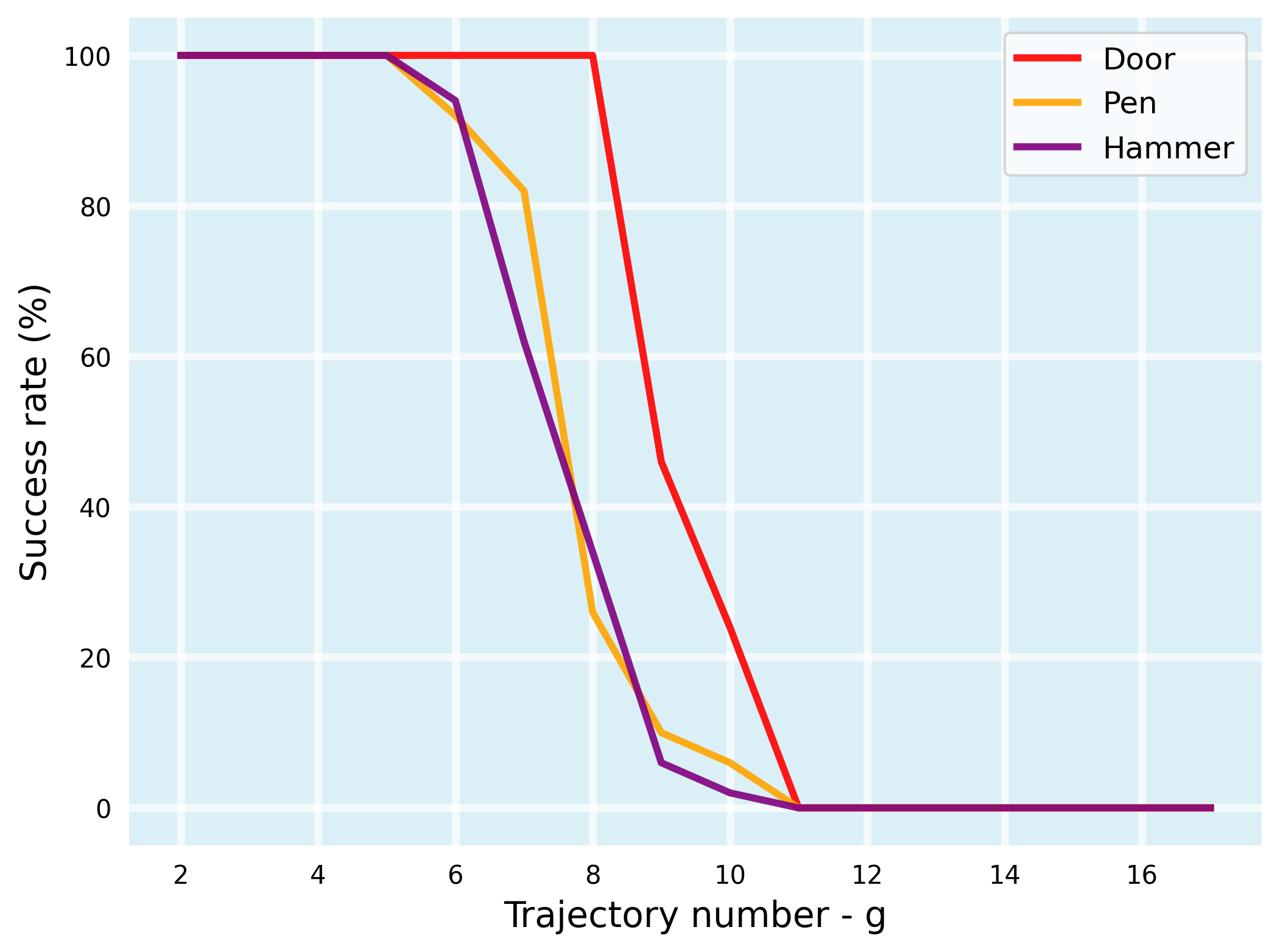}}
        \subfigure[Locomotion]{\includegraphics[width=0.3\textwidth]{ 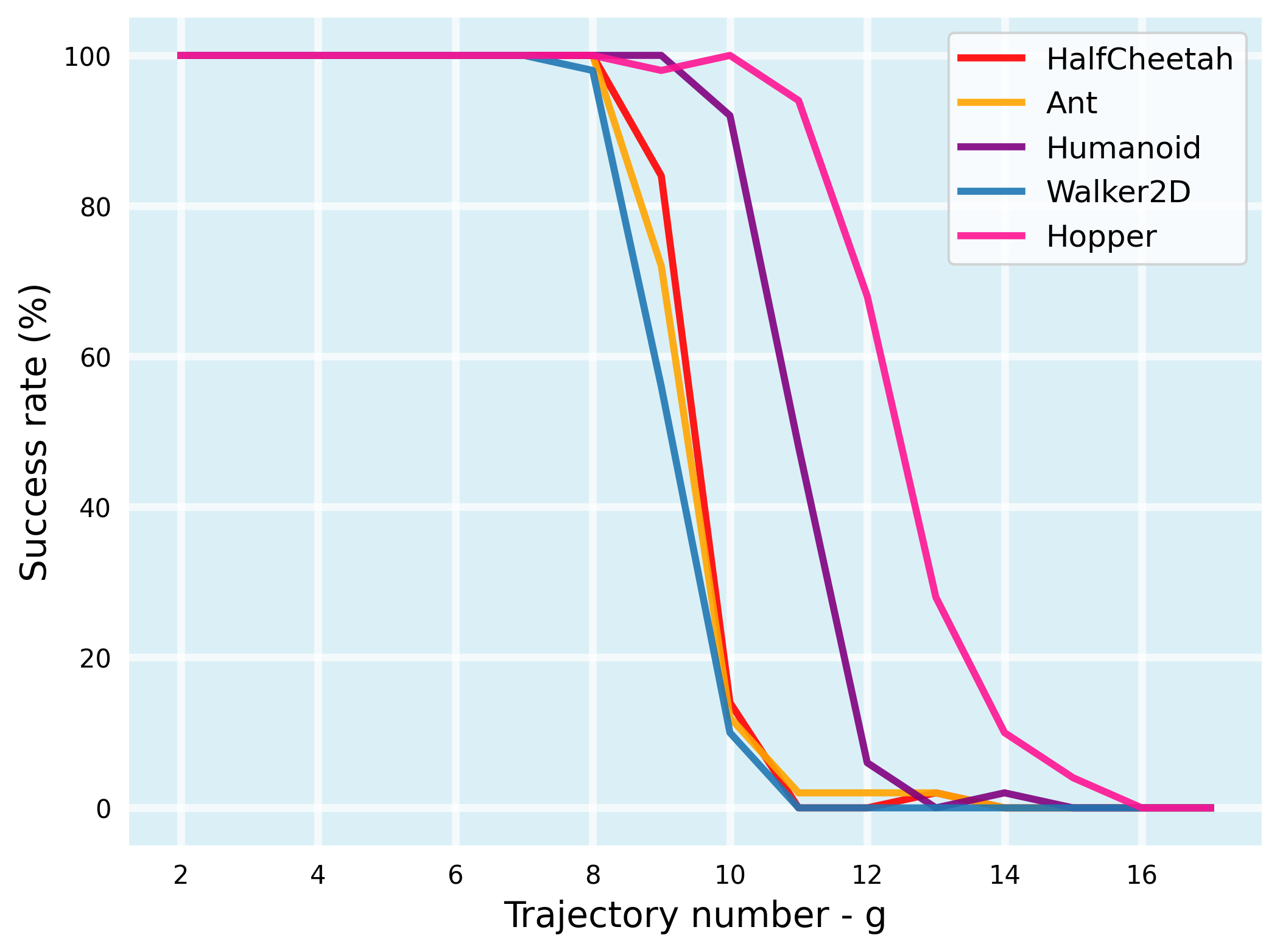}}
        \subfigure[TriFinger]{\includegraphics[width=0.3\textwidth]{ 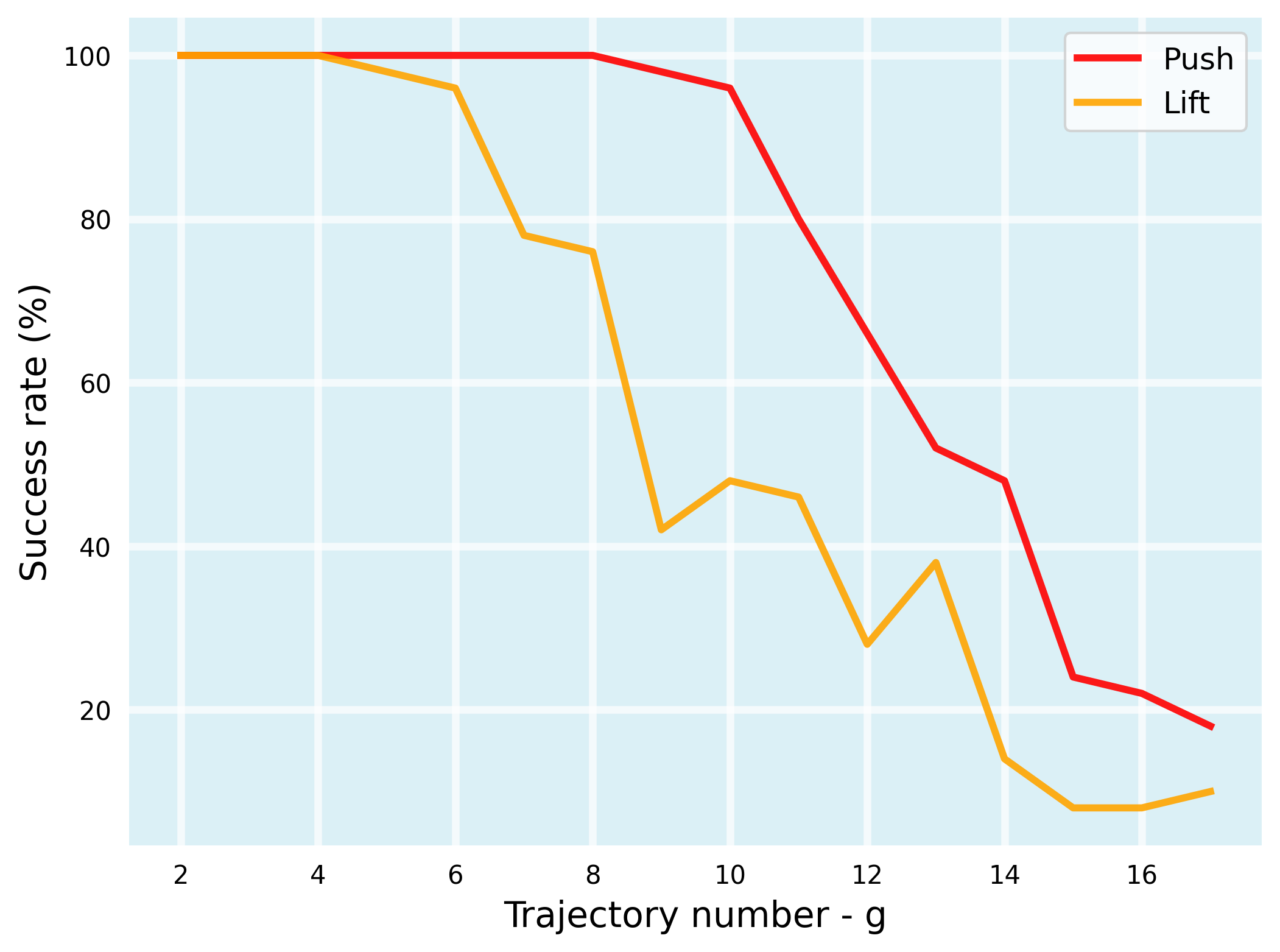}}
    \caption{The success rate of MCS in find a subset that is uni-behavioral, explored across a range of $g$ value choices.}
    \label{fig:assum51-multi-k}  
\end{figure*}

We employed the Monte Carlo search (MCS) method, as outlined in Section \ref{subsec:extract-seed-set} of the main text, to identify the subset $G$ with the highest density (expected) in the multi-behavior dataset. We then assessed whether this subset $G$ is uni-behavior. To account for the randomness of MCS, we conducted each search experiment $100$ times, counting how many times $G$ successfully exhibited uni-behavior. We varied the number of trajectories within $G$ (denoted as $g$), recorded the success rate for each $g$ value, and subsequently generated Figure \ref{fig:assum51-multi-k}. It can be observed that when the value of $g$ is relatively small, i.e., $g \le 6$, the vast majority of the $G$ obtained through MCS exhibit uni-behavior. Considering that having a larger number of trajectories in $G$ is more beneficial for subsequent operations, we opted to set $g$ to $6$ in our experiments.

\section{Utilizing the elbow method to determine the number of clusters for K-means} \label{append:kmeans-elbow}
This part of the experiment aims to support the viewpoint we proposed in the Section \ref{sec:clu-main}, that the elbow method is not applicable for clustering tasks on multi-behavior datasets when using the K-means. The elbow method is a technique to determine the number of clusters by identifying the “elbow” in the plot of the sum of squared errors (SSE) versus assumed cluster number. SSE refers to the sum of squares of the distances of each point to its nearest cluster center in K-means clustering. As the number of clusters increases, the SSE will decrease. However, the “elbow” point in the SSE curve, where the rate of decline in SSE significantly slows down, is usually considered as an indicator that the optimal number of clusters has been found and that adding further clusters will only serve to divide true clusters. 

\begin{figure*}[htbp]
    \centering
        \subfigure[Hand]{\includegraphics[width=0.3\textwidth]{ 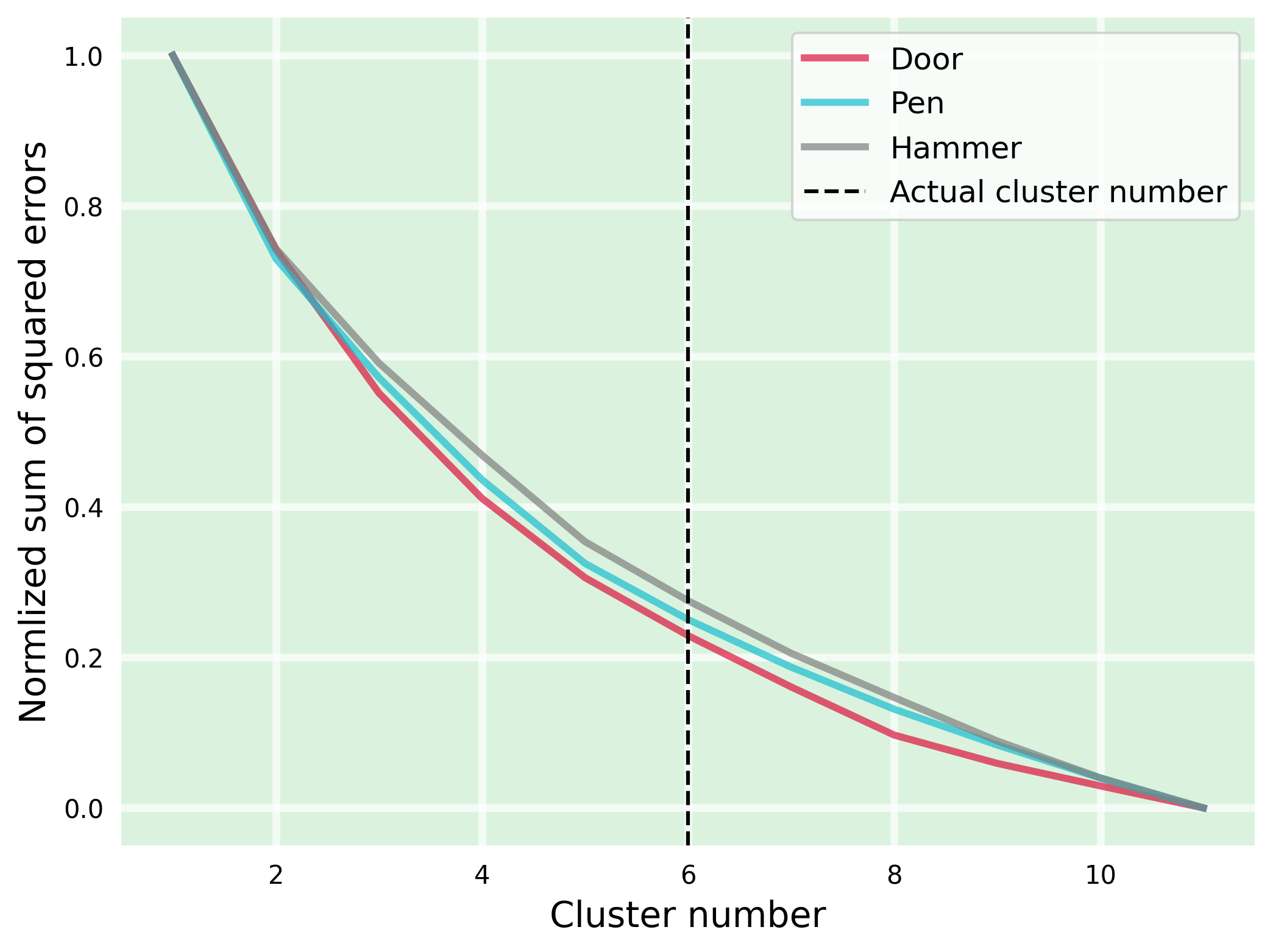}}
        \subfigure[Locomotion]{\includegraphics[width=0.3\textwidth]{ 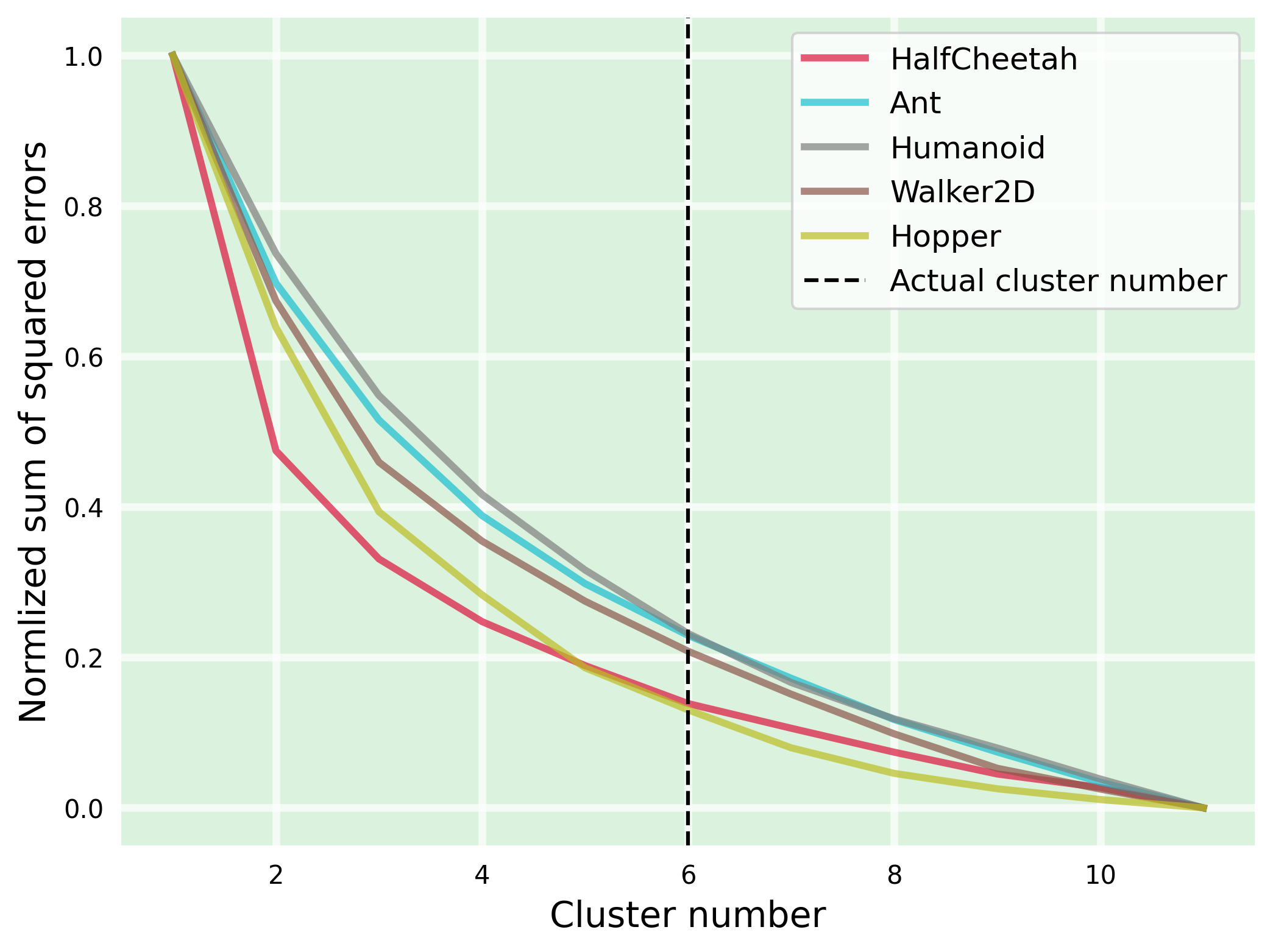}}
        \subfigure[TriFinger]{\includegraphics[width=0.3\textwidth]{ 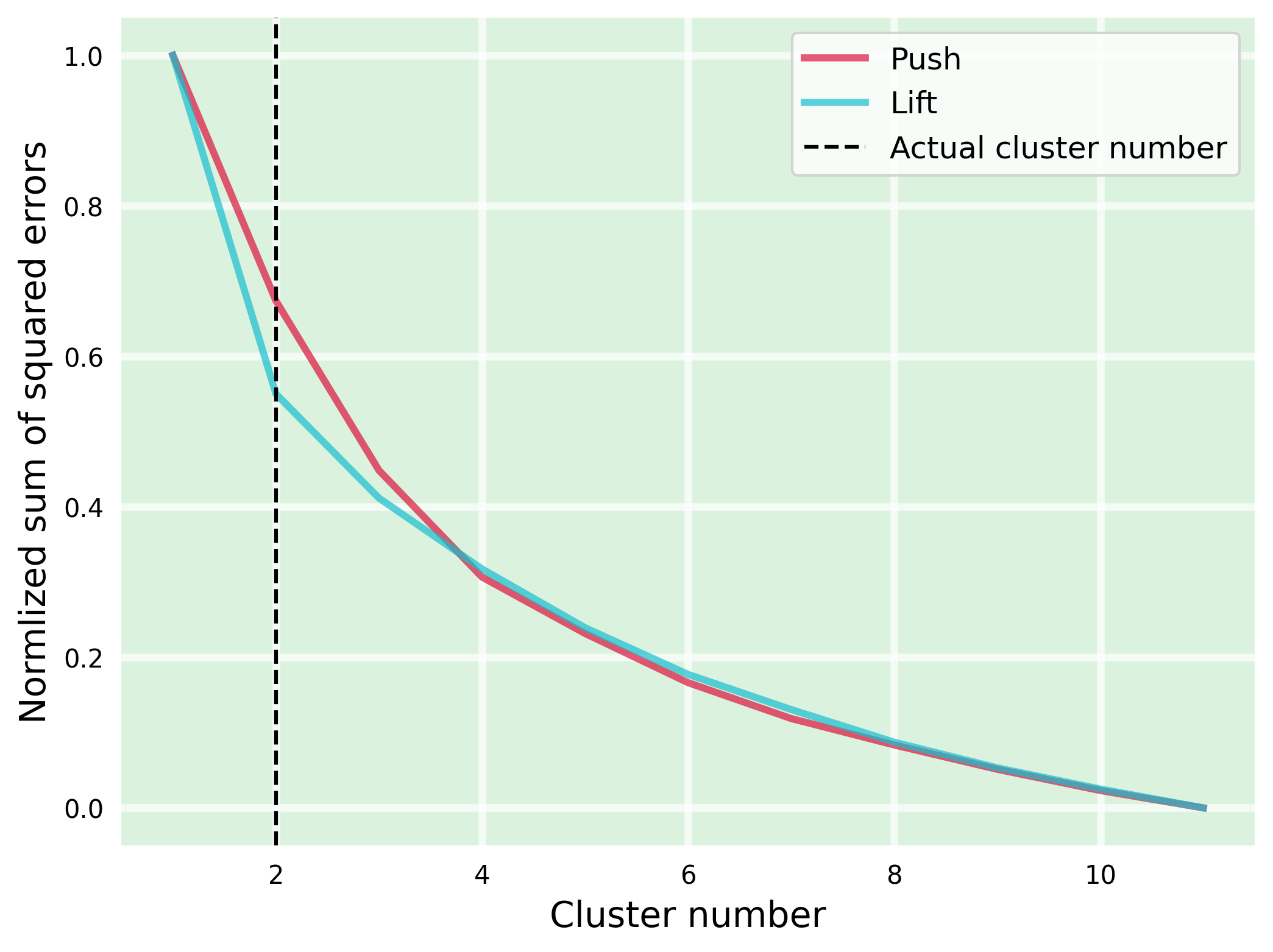}}
    \caption{The plot displays the SSE calculated across a range of cluster numbers using K-means on the TAAT data. Each colored line indicates an instance of each dataset, and the black dashed line indicates the ground truth values of cluster numbers.}
    \label{fig:kmeans-elbow}  
\end{figure*}

In this experiment, we apply K-means to the TAAT data and set the cluster number to various values. Subsequently, we calculate the SSE for each clustering result under each cluster number and then plot the results in Figure \ref{fig:kmeans-elbow}. However, for all the multi-behavior datasets examined, no distinct elbow shape appeared when the number of clusters was the actual values. This indicates that the elbow method is not effective in finding the optimal number of clusters for such datasets, likely due to significant overlap in the cluster distributions.

\section{Detailed algorithm} \label{append:algorithm}
We detail the iterative process of our behaviour-aware clustering approach in Algorithm \ref{alg:clu}.
\begin{algorithm*}[!htbp]
   \caption{Behaviour-aware clustering algorithm}
   \label{alg:clu}
\begin{algorithmic}
\STATE {\bfseries Input:} Multi-behaviour dataset $\mathcal{D}$
   \STATE Initialise a buffer $\mathcal{D}_{out}$ to store the clustered subsets;
   \WHILE{not the last cluster}
       \STATE Get an uni-behaviour seed dataset $\mathcal{D}_{seed}$;
       \WHILE{$\mathcal{D}_{seed}$ not converged}
           \STATE Initialize a classifier ${\mathcal{F}_{\theta}}$ with the neural network parameters $\theta$;
           \FOR{$epoch=1$ {\bfseries to} $epochs$}
                \STATE Update $\theta$ by minimizing the loss in Equation~\ref{eq:cross-entropy-loss};
           \ENDFOR
           \STATE Apply the trained ${\mathcal{F}_{\theta}}$ on $\mathcal{D}$ to update the membership of $\mathcal{D}_{seed}$;
       \ENDWHILE
       \STATE $\mathcal{D}_{out}$ $=$ $\mathcal{D}_{out}$ $+$ $\mathcal{D}_{seed}$;
       \STATE Determine if the current cluster is the final one;
   \ENDWHILE
\STATE {\bfseries Output:} $\mathcal{D}_{out}$; it contains multiple uni-behavior datasets
\end{algorithmic}
\end{algorithm*}

\section{Supplementary examples for understanding the positive-unlabelled filter} \label{append:sup-pu}
\subsection{Adaptive threshold for converting continuous probabilities into Boolean values} \label{append:subsec-adaptive-th}
\begin{figure}[ht]
\vskip 0.2in
\begin{center}
    \centerline{\includegraphics[width=0.95\linewidth]{ 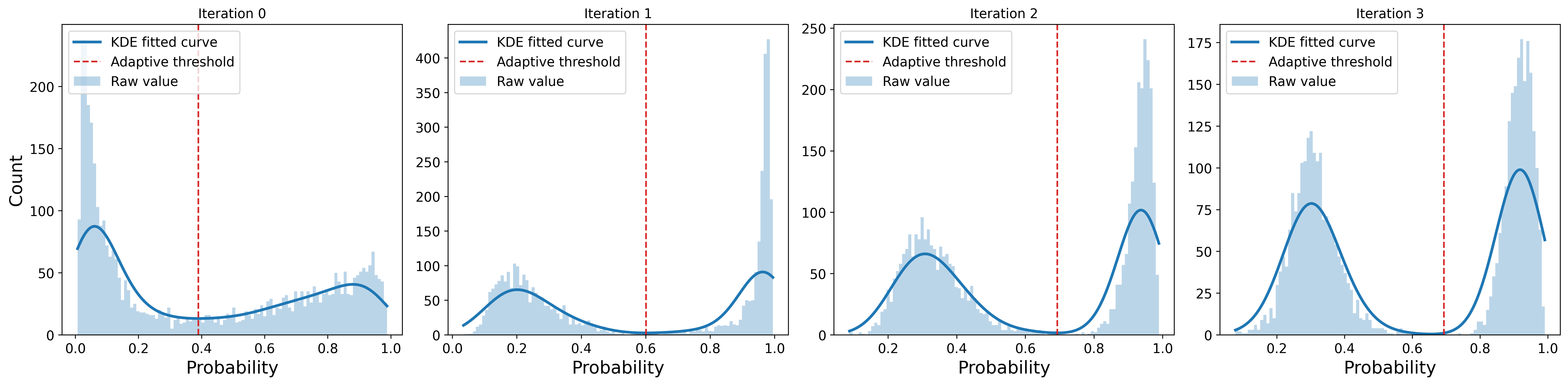}}
    \caption{An example (the push task within the Trifinger domain) illustrates the selection of thresholds for converting continuous probabilities into Boolean values across several consecutive semi-supervised iterations of the PU filter. The red dashed line indicates the selected threshold.}
\label{fig:adap-th}  
\end{center}
\vskip -0.2in
\end{figure}

This section serves as supplementary material to Section \ref{subsubsec:use-pu-filter}, with the aim of visually demonstrating how our positive-unlabelled (PU) filter automatically determines the threshold for converting continuous probability values into discrete Boolean values. An example is shown in Figure \ref{fig:adap-th}: We first draw a histogram based on the probability values output by the PU filter and then use kernel density estimation (KDE) to approximate the distribution of these data. While the original work employs polynomial methods for fitting \cite{pubc}, we have found that KDE offers more precise and smoother results. By analyzing the KDE function, we identify its local minima and select the probability value at the largest local minima as the threshold. With this threshold, the PU filter can effectively extract data that are expected to mostly contain uni-behavior data, sharing the same behavioural pattern as the seed data.

\subsection{Determining whether it is the last cluster} \label{append:subsec-end-of-clu}
\begin{figure}[ht]
\vskip 0.2in
\begin{center}
    \centerline{\includegraphics[width=0.7125\linewidth]{ 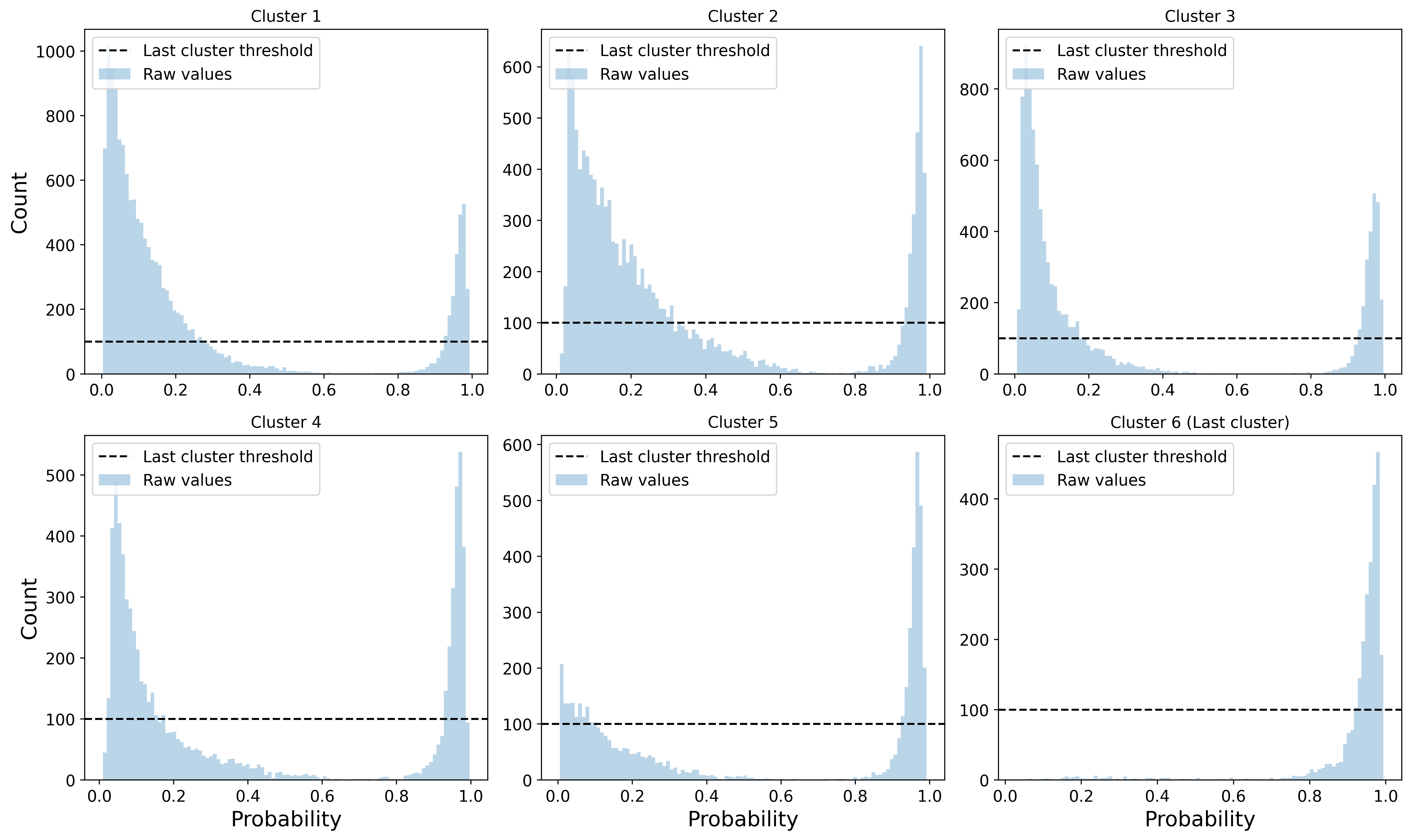}}
    \caption{An example (the hammer task within the robotic hand domain) illustrates the determination of whether the current cluster should be the final one, where the black dashed line represents the specified threshold.}
\label{fig:check-last-cluster}  
\end{center}
\vskip -0.2in
\end{figure}

This section provides a visual demonstration of how our method determines whether the current cluster is the last one, complementing Section \ref{subsec:check-last-cluster}. We illustrate this with an example in Figure \ref{fig:check-last-cluster}, where each subplot represents the probability distribution histograms obtained in the final round of the PU filter for each cluster. In the preceding iterations, two distinct peaks are shown, with the larger peak representing data having the same behavior pattern as in the seed, and the lower one comprising other data.  As the clusters progress through iterations, the lower probability distribution gradually diminishes in data count until it is almost non-existent in the final cluster. Therefore, we terminate the clusting process when the data count in the lower probability distribution falls below a certain threshold (specifically, it ranges from 0.1\% to 2\% of the entire dataset size in our work), or when the lower probability peak cannot be detected at all.

\section{Advantages of using a clustered subset for offline policy learning} \label{appendix:advantages-using-clustering-for-policy}
We separately train policies using each of the clustered uni-behavior subsets, then select the best-performing policy among them to compare it with the policy trained on the raw multi-behavior dataset. We report the results in Table \ref{table:policy-results-use-clustered}. It should be noted that only tasks within the Locomotion and Robotic Hand domains are included and the tasks within Trifinger domain are not included. This is because the Trifinger dataset was not collected by us, and we do not have direct access to these tasks for evaluation. Each Trifinger dataset was collected using two policies: one being an expert policy, and the other a weaker policy. Previous studies have shown that policies trained using expert-generated training data subsets consistently outperformed policies trained using multi-behavior datasets in various testing scenarios \cite{benchmarking-trifinger}. Here, our clustering algorithm achieves near-perfect performance, leading to the expectation that the best policies trained from our clustered subsets should exhibit performance similar to the policies trained on expert datasets in \citet{benchmarking-trifinger}. Therefore, interested readers can find additional results and comprehensive analysis in \citet{benchmarking-trifinger}.

\begin{table}[!htbp]
\caption{The normalized evaluation results of different algorithms on the uni-behavior clustered subsets and the raw multi-behavior dataset. Each evaluation result is based on $20$ evaluation episodes and $3$ different seeds. "*" denotes the evaluation results of the best-performing policy among all policies trained from each clustered subset.}
\centering
\begin{tabular}{l||cc|cc|cc|cc}
\multirow{2}{*}{} & \multicolumn{2}{c|}{BC} & \multicolumn{2}{c|}{CRR} & \multicolumn{2}{c|}{TD3PlusBC} & \multicolumn{2}{c}{IQL} \\ \cline{2-9} 
                 & Uni*  & Multi & Uni*  & Multi & Uni*  & Multi & Uni*  & Multi \\ \hline\hline
Loco- Ant        & 0.805 & 0.458 & 0.811 & 0.509 & 0.789 & 0.773 & 0.802 & 0.735 \\
Loco-Halfcheetah & 0.937 & 0.531 & 0.875 & 0.856 & 0.813 & 0.829 & 0.853 & 0.533 \\
Loco-Hopper      & 0.878 & 0.500 & 0.835 & 0.308 & 0.975 & 0.498 & 0.851 & 0.421 \\
Loco-Walker2d    & 0.958 & 0.829 & 0.921 & 0.189 & 0.925 & 0.877 & 0.845 & 0.553 \\
Loco-Humanoid    & 0.901 & 0.758 & 0.487 & 0.566 & 0.588 & 0.698 & 0.921 & 0.766 \\ \hline
Loco-Average     & 0.896 & 0.615 & 0.786 & 0.486 & 0.818 & 0.735 & 0.854 & 0.602 \\ \hline\hline
Hand-Hammer      & 0.872 & 0.559 & 0.664 & 0.613 & 0.881 & 0.356 & 0.939 & 0.781 \\
Hand-Door        & 0.969 & 0.662 & 0.887 & 0.295 & 0.511 & 0.293 & 0.918 & 0.774 \\
Hand-Pen         & 0.886 & 0.696 & 0.896 & 0.816 & 0.948 & 0.958 & 0.822 & 0.655 \\ \hline
Hand-Average     & 0.909 & 0.639 & 0.816 & 0.575 & 0.780 & 0.536 & 0.893 & 0.737 \\ \hline\hline
Average           & \textbf{0.902}  & 0.627 & \textbf{0.801}  & 0.530  & \textbf{0.799}   & 0.635   & \textbf{0.874}  & 0.669
\end{tabular}
\label{table:policy-results-use-clustered}
\end{table}

In our work, the best-performing policy trained from the uni-behavior subsets outperforms that trained on the multi-behavior dataset, despite the latter having a six-fold data advantage in terms of available training data. However, it is important to emphasize that this is contingent upon the clustered subset not being too small, as training a policy becomes challenging when the data volume and diversity are extremely limited.

\section{Substitute for the formulation of the TAAT}
We present a formulation for the TAAT, using the geometric mean as a substitute for Equation \ref{eq: action-rep}:
\begin{equation}
\label{eq: geo}
\bm{\overline{\tau^{\mathcal{A}}}}   = \left[\exp(\textstyle \frac{1}{T} \sum_{t=1}^{T} \log(a_{t,1})), \exp(\textstyle \frac{1}{T}\sum_{t=1}^{T} \log(a_{t,2})), ..., \exp(\textstyle \frac{1}{T}\sum_{t=1}^{T} \log(a_{t,i}))\right].
\end{equation}
Although experimental tests indicate that its performance is similar to that of Equation \ref{eq: action-rep} when applied to our dataset and involves relatively more complex computation than Equation \ref{eq: action-rep}, we believe it may be advantageous in scenarios with highly fluctuating action trajectories or data containing glitches, such as unexpected sensory glitches occasionally encountered in real-world robotic systems. This advantage arises from the noise-resistant characteristics inherent in the geometric mean.

\section{Composition of datasets} \label{append:dset-compo}
We present the composition of each uni-behaviour dataset used in this work in Tables \ref{table:policy-dset-compo} and \ref{table:clu-dset-compo}. These tables detail: (1) the learning algorithm used to train the policy for data collection, among PPO \cite{ppo}, SAC \cite{sac} and TD3 \cite{td3}; (2) the exploration noise introduced into the datasets; (3) the number of transitions in each dataset; and (4) the performance of the policy used for collecting each dataset. The datasets listed in Table \ref{table:policy-dset-compo} are utilized for generating the results in Appendix \ref{append:multi-vs-uni-results}, and Table \ref{table:clu-dset-compo} outlines the main clustered datasets used in this paper.

\begin{table}[ht]
\centering
\caption{The composition of datasets used for generating the results in Appendix \ref{append:multi-vs-uni-results}.}
\label{table:policy-dset-compo}
\begin{tabular}{l||c|r|r|rrr}
\multirow{2}{*}{} & \multirow{2}{*}{Algorithm} & \multirow{2}{*}{\shortstack{Exploration\\noise}} & \multirow{2}{*}{\shortstack{Transition\\[-0.15em]count}} & \multicolumn{3}{c}{Mean return} \\ \cline{5-7} 
                 &     &      &         & \multicolumn{1}{c}{Set 1} & \multicolumn{1}{c}{Set 2} & \multicolumn{1}{c}{Set 3}  \\ \hline\hline
Loco-Ant        & SAC & $\mathcal{N}(0, 0.2)$  & $1.5\times10^{6}$ & 6174.25  & 6693.88  & 6615.37  \\
Loco-Hopper      & TD3 & $\mathcal{N}(0, 0.2)$  & $1.5\times10^{6}$ & 3719.61  & 3776.64  & 3610.47  \\
Loco-Walker2d    & SAC & $\mathcal{N}(0, 0.15)$ & $1.5\times10^{6}$ & 6594.58  & 6401.43  & 6178.12  \\
Loco-Halfcheetah & SAC & $\mathcal{N}(0, 0.15)$ & $1.5\times10^{6}$ & 13390.75 & 13056.99 & 13087.77
\end{tabular}
\end{table}

\begin{table}[ht]
\centering
\caption{The composition of the clustering datasets.}
\label{table:clu-dset-compo}
\resizebox{\textwidth}{!}{
\begin{tabular}{l||c|r|r|rrrrrr}
\multirow{2}{*}{} & \multirow{2}{*}{Algorithm} & \multirow{2}{*}{\shortstack{Exploration\\noise}} & \multirow{2}{*}{\shortstack{Transition\\[-0.15em]count}} & \multicolumn{6}{c}{Mean return} \\ \cline{5-10}
                 &     &      &         & \multicolumn{1}{c}{Set 1} & \multicolumn{1}{c}{Set 2} & \multicolumn{1}{c}{Set 3} & \multicolumn{1}{c}{Set 4} & \multicolumn{1}{c}{Set 5} & \multicolumn{1}{c}{Set 6} \\ \hline\hline
Loco-Ant        & SAC & $\mathcal{N}(0, 0.2)$  & $3\times10^{6}$ & 6897.74  & 7166.93  & 4686.98 & 4811.98 & 3274.83 & 3225.65 \\
Loco-Hopper      & TD3 & $\mathcal{N}(0, 0.2)$  & $3\times10^{6}$ & 3719.61  & 3776.64  & 2705.04 & 2745.87 & 1270.03 & 1286.49 \\
Loco-Humanoid    & SAC & $\mathcal{N}(0, 0.2)$  & $3\times10^{6}$ & 7377.83  & 7466.53  & 5511.97 & 5344.22 & 2500.22 & 2884.44 \\
Loco-Halfcheetah & SAC & $\mathcal{N}(0, 0.15)$ & $3\times10^{6}$ & 13390.75 & 13056.99 & 7931.48 & 8247.26 & 4052.93 & 4157.20  \\
Loco-Walker2d    & SAC & $\mathcal{N}(0, 0.15)$ & $3\times10^{6}$ & 6499.63  & 6401.43  & 4585.49 & 4641.82 & 2714.81 & 3211.37 \\
Hand-Hammer      & SAC & $\mathcal{N}(0, 0.2)$  & $3\times10^{6}$ & 16419.66 & 17937.65 & 2943.85 & 3801.33 & -250.77 & -126.30  \\
Hand-Pen         & SAC & $\mathcal{N}(0, 0.2)$  & $3\times10^{6}$ & 4430.05  & 4723.89  & 3554.91 & 3269.03 & 2061.42 & 1976.53 \\
Hand-Door        & TD3 & $\mathcal{N}(0, 0.2)$  & $3\times10^{6}$ & 2849.22  & 3120.38  & 1443.70  & 1219.78 & -55.22  & -14.14  \\
Trifinger-Push   & PPO & -    & $2.8\times10^{6}$ & 660.14   & 197.95   & -       & -       & -       & -       \\
Trifinger-Lift   & PPO & -    & $3.6\times10^{6}$ & 1063.99  & 638.29   & -       & -       & -       & -      
\end{tabular}
}
\end{table}

\section{Hyperparameters} \label{append-hyper}
Our code and datasets have all been open-sourced on our website \url{https://github.com/wq13552463699/Clustering-For-Offline-RL-IL}. For training the policies included in our paper, we utilized the available open-source library d3rlpy \cite{d3rlpy} using the recommended hyperparameters. The key hyperparameters used for the clustering algorithm of each dataset are introduced on our website.

\end{document}